\documentclass[journal,compsoc]{IEEEtran}
\usepackage{algorithm,algorithmic}
\usepackage{amsmath}
\usepackage{anyfontsize}
\usepackage{hyperref}
\hypersetup{colorlinks = true,
            linkcolor = blue,
            urlcolor = blue, 
            citecolor=red}
\renewcommand\footnoterule{\kern-3pt \hrule width 3.5in \kern 2.6pt}
\newcommand\Tstrut{\rule{0pt}{2.3ex}}       
\makeatletter
\def\BState{\State\hskip-\ALG@thistlm}
\makeatother

\ifCLASSINFOpdf
  \usepackage[pdftex]{graphicx}
\else
\fi
\ifCLASSOPTIONcompsoc
\usepackage[caption=false,font=normalsize,labelfont=sf,textfont=sf]{subfig}
\else
  \usepackage[caption=false,font=footnotesize]{subfig}
\fi
\usepackage[pscoord]{eso-pic}
\newcommand{\placetextbox}[3]{
	\setbox0=\hbox{#3}
	\AddToShipoutPictureFG{ \put(\LenToUnit{#1\paperwidth},\LenToUnit{#2\paperheight}){\vtop{{\null}\makebox[0pt][c]{#3}}}
	}
}
\placetextbox{.5}{0.997}
{\fontsize{6}{6}\selectfont
	This article has been accepted for publication in a future issue of this journal, but has not been fully edited. Content may change prior to final publication. Citation information: DOI 10.1109/TCBB.2019.2911609, IEEE/ACM
}
\placetextbox{.5}{0.988}
{\fontsize{6}{6}\selectfont
 Transactions on Computational Biology and Bioinformatics
}

\placetextbox{.5}{0.035}
{\fontsize{6}{6}\selectfont
	Copyright (c) 2019 IEEE. Personal use is permitted, but republication/redistribution requires IEEE permission. See http://www.ieee.org/publications\_standards/publications/rights/index.html for more information.
}

\begin{document}
\title{Deep Robust Framework for Protein Function Prediction using Variable-Length Protein Sequences}
%
%
%

\author{Ashish~Ranjan,
		Md~Shah~Fahad,
        David~Fern$\acute{\text{a}}$ndez-Baca,
        Akshay~Deepak,
        and~Sudhakar~Tripathi
        
\IEEEcompsocitemizethanks{
\IEEEcompsocthanksitem Ashish Ranjan (corresponding author), Md Shah Fahad and Akshay Deepak are with the Department of Computer Science \& Engineering, National Institute of Technology Patna, India. E-mail: \{ashish.cse16, shah.cse16, akshayd\}@nitp.ac.in.
\IEEEcompsocthanksitem David Fern$\acute{\text{a}}$ndez-Baca is with the Department of Computer Science, Iowa State University, USA. E-mail: fernande@iastate.edu%
\IEEEcompsocthanksitem Sudhakar Tripathi is with the Department of Information Technology, REC Ambedkar Nagar, India.}
}

\IEEEtitleabstractindextext{
\begin{abstract}
The order of amino acids in a protein sequence enables the protein to acquire a conformation suitable for performing functions, thereby motivating the need to analyse these sequences for predicting functions. Although machine learning based approaches are fast compared to methods using BLAST, FASTA, etc., they fail to perform well for long protein sequences (with more than 300 amino acids). In this paper, we introduce a novel method for construction of two separate feature sets for protein  using bi-directional long short-term memory network based on the analysis of fixed 1) single-sized segments and 2) multi-sized segments. The model trained on the proposed feature set based on multi-sized segments is combined with the model trained using state-of-the-art Multi-label Linear Discriminant Analysis (MLDA) features to further improve the accuracy. Extensive evaluations using separate datasets for biological processes and molecular functions demonstrate not only improved results for long sequences, but also significantly improve the overall accuracy over state-of-the-art method. The single-sized approach produces an improvement of +3.37\% for biological processes and +5.48\% for molecular functions over the MLDA based classifier. The corresponding numbers for multi-sized approach are +5.38\% and +8.00\%. Combining the two models, the accuracy further improves to +7.41\% and +9.21\% respectively.
\end{abstract}

}


\maketitle


%
\IEEEpeerreviewmaketitle

\section{Introduction}
\IEEEPARstart{P}{roteins} are among the most elementary yet complex functional molecules found in all living organisms. Their presence enables various molecular and biological processes that are essential for smooth operation of different biological components in an organism. Among other things, they act as catalysis and transporting agents, signal molecules and form structural support for an organism \cite{Pandey}. Consequently, estimation of protein functions is important for decoding the underlying mechanism of various complex biological processes of an organism. This is also helpful in the treatment of certain diseases and development of new drugs. 

Although laboratory experimental procedures are reliable for estimation of protein functions, they are slow and expensive \cite{Pandey}. To overcome the limitations of experimental procedures, researchers have been developing new computational approaches to protein function prediction. These include analyzing protein sequences \cite{PSSM-function,ProLanGo,DeepGo,FANNGO}, protein structures \cite{structure1,structure2,Co-factor} and protein-protein interaction (PPI) networks  \cite{PPI-Function1,PPI-Function2}. Other approaches are hybrid in the sense that they combine two or more of these three approaches  \cite{DeepGo,Ms-knn,Inga,MLDA_protFun}. However, the availability of protein structures and protein interaction networks is much more limited (restricted to a few organisms only) than for protein sequences. The recent successful advent of next generation sequencing technologies \cite{NextGenSeq} have led to a deluge of protein sequences. This has elicited computational biologists to identify the structural \cite{structure_lstm,Struc-seq} and functional \cite{ProLanGo,DeepGo} behaviour of uncharacterized proteins using characterized ones by analyzing their sequential data. In this context, analyzing and decoding hidden patterns from protein sequences is a crucial task for understanding the functional roles of a protein.

Function characterization of a protein based upon analysis of its amino acid sequence is a complex procedure. In \cite{Pandey}, existing procedures were put into three categories as: \textit{Homology based approaches}, \textit{Subsequence based approaches} and \textit{Feature based approaches}. In \textit{Homology based approaches}, annotations depend upon similarity score with known protein sequence using sequence alignment techniques such as FASTA \cite{Fasta} and BLAST \cite{Blast,Psi-BLAST}. These approaches usually correlate sequence similarity with functional similarity. But, studies highlight that such rationale based on similarity is a weak hypothesis \cite{structure1}. \textit{Subsequence based approaches} search for hidden recurrent patterns or segments in a protein sequence such as motifs \cite{Seq_motif,Seq_motif1} and/or functional domain. However, identifying such latent pattern given variable-length protein sequences is still a bottleneck problem. \textit{Feature based approaches} transform raw protein sequences into discriminative features for efficient characterization of new proteins based on machine learning techniques. These approaches frequently use \textit{n}-mers for the construction of the sequence vector for a protein \cite{MLDA_protFun,PPI}. Apart from \textit{n}-mers, PSSM (Position Specific Scoring Matrix) features are also used for functional annotation \cite{PSSM-function}.

Recently, several neural network based techniques have been introduced for function characterization based on the analysis of entire protein sequences. Cao \textit{et al.} \cite{ProLanGo} proposed a Neural Machine Translation (NMT) based method to translate a protein sequence into possible $GO$-terms by treating both sequence and $GO$-terms as a language. In another approach, Kulmanov \textit{et al.} \cite{DeepGo} introduced a Convolutional Neural Network (CNN) based method and used it in combination with a hierarchy of neural networks for each $GO$-term to predict the function(s) of query protein sequence. Clark \textit{et al.} \cite{FANNGO} formulated a protein vector based on \textit{i}-score for each $GO$-term and used neural network ensemble for function prediction.

All methods listed above have made great contributions towards
protein function prediction based on protein sequences.
However, there still exist few shortcomings that need attention from machine learning prospective, thereby providing motivation for our work as follows:
\begin{itemize}
	\item \textbf{Dealing with variation in sequence lengths.} Protein sequences vary in length and can be partitioned into \textit{short}, \textit{medium} and \textit{long} sequences.  Generally, protein sequence datasets have a mix of short, medium and long sequences. While several approaches have been proposed to transform varied length protein sequences to equal-sized protein vectors (e.g., \cite{MLDA_protFun, SGT}), the proposed methods often fail to perform for \textit{long} protein sequences. One of the main reasons for this is that the majority of protein sequences are in short-to-medium range, resulting in an uneven distribution with respect to sequence lengths as shown in Fig. \ref{Seq_dist}.\vspace{2mm}

	\item \textbf{Efficient feature vector representation.} Every sequence is assumed to have \textit{conserved} regions, which are responsible for functional classification of sequences, and \textit{non-conserved} regions, which don't contribute to functional classification. In this sense, the \textit{non-conserved} regions are also seen as \textit{noise}. As the length of sequences increase, the \textit{noisy} regions tend to out-weight \textit{conserved} regions in terms of relative lengths. As a result, the existing vector representations of protein sequences tend to get affected by \textit{non-conserved} regions resulting in low performance, especially for longer sequences. Further, detailed discussion on this aspect is provided in Sec. \ref{seq_seg}.\vspace{2mm}
	
	\item \textbf{Performance Improvement}: To improve upon the overall prediction accuracy compared to existing methods in case of varied lengths.
\end{itemize}

%
%

Based on the above discussion, the objective of our work is:
\textit{To develop a multi-label protein function classification framework that gives superior performance with respect to existing approaches and  produces consistent results for sequences of varying lengths, especially for long sequences.} 

Protein functions are represented as \textit{Gene Ontology} (GO) \cite{GO} terms. GO produces unified vocabulary for gene products across three domains: \textit{Cellular Component} (CC), \textit{Molecular Function} (MF), and \textit{Biological Process} (BP). To the best of our knowledge, no work has been done to specifically handle \textit{long} sequences while preserving the prediction accuracy of \textit{short} and \textit{medium} sequences. In this regard, the contributions of this paper are as follows:

\begin{figure}[!t]
	\centering
	\subfloat[Biological Process]{\includegraphics[width=3.4in]{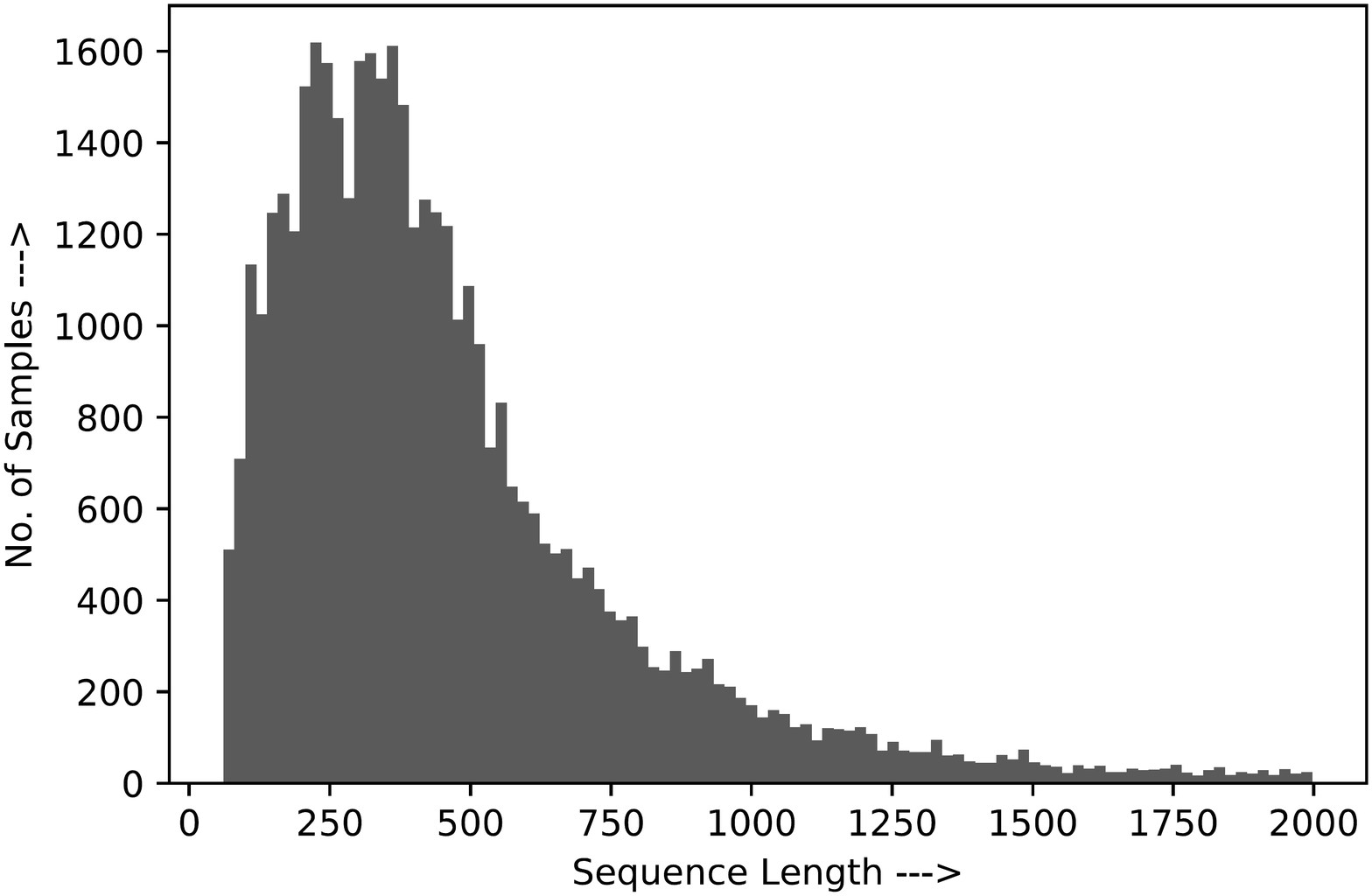}\label{seq_dist_BP}}
	
	\subfloat[Molecular Function]{\includegraphics[width=3.4in]{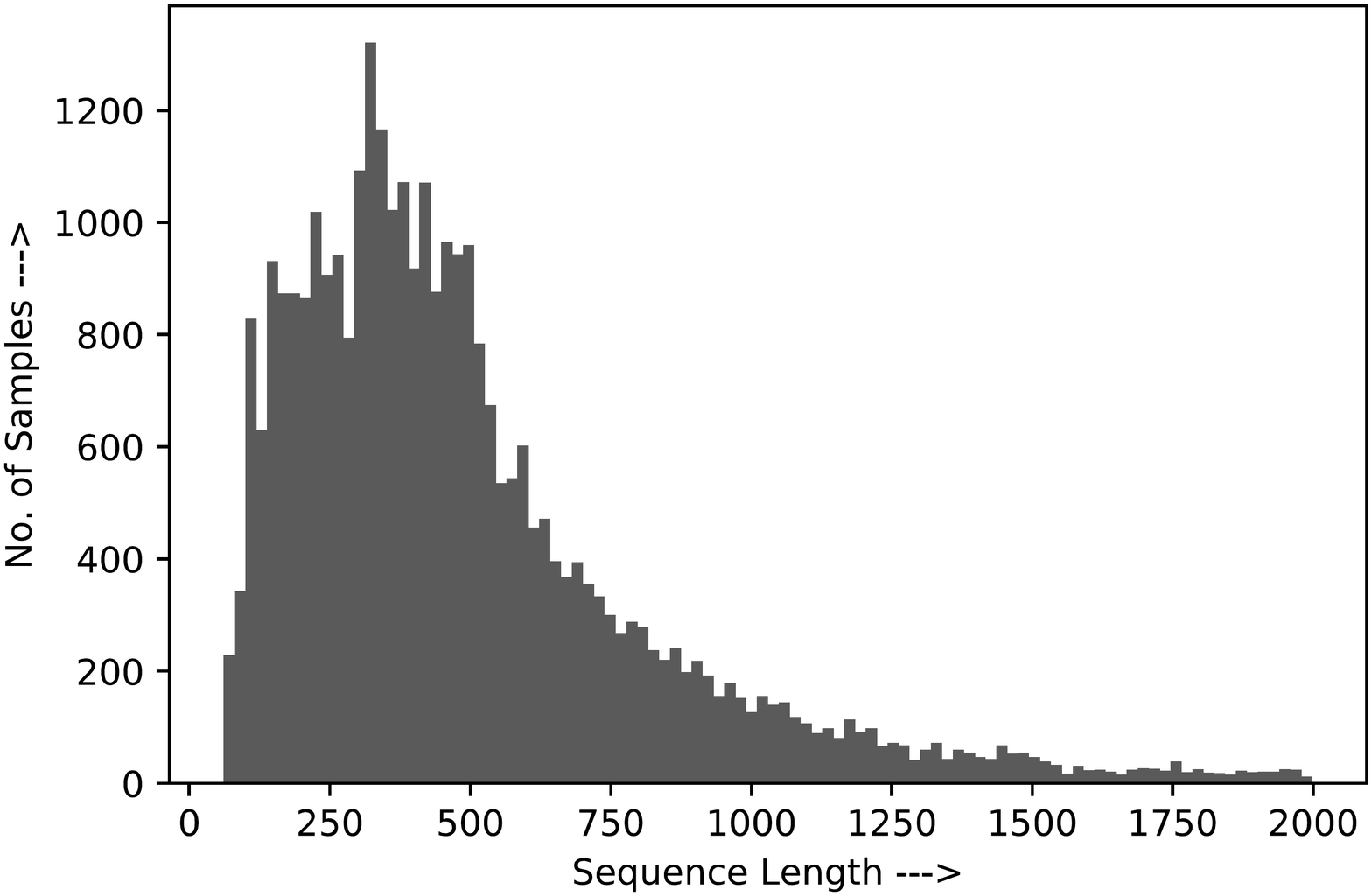}\label{seq_dist_MF}}
	
	\caption{Length distribution of protein sequences used for experimental evaluation. The sequences were obtained from UniProtKB database (\url{https://www.uniprot.org}) \cite{uniprot}. Distribution of protein sequences for biological process prediction is shown in (\ref{seq_dist_BP}), while (\ref{seq_dist_MF}) shows the corresponding graph for molecular function prediction.}
	\label{Seq_dist}
\end{figure}

\begin{itemize}
\item \textbf{Efficient feature vector representation}: A novel approach to protein vector construction \textit{ProtVecGen}, based on the proposed segmentation technique and bi-directional LSTM networks \cite{BiLstm} is introduced. The proposed technique produces a fixed-length vector for a protein sequence that is robust for function prediction with respect to high variation in sequence lengths.\vspace{2mm}

\item \textbf{Multi-sized segmentation}: Protein vector construction is further improved by combining multiple \textit{ProtVecGen} features based on different segment sizes to yield a more discriminative set of features called \textit{ProtVecGen-Plus}. This significantly improved the performance of the proposed framework compared to \textit{ProtVecGen}.\vspace{2mm}

\item \textbf{Hybrid approach}: The classification model based on \textit{ProtVecGen-Plus} features is combined with another model based on MLDA features \cite{MLDA_protFun} to produce even better results.\vspace{2mm}

\item \textbf{Superior performance}: The proposed features are evaluated using protein sequences from UniProtKB dataset \cite{uniprot} with 58310 protein sequences  (having 295 distinct \textit{GO}-terms) for BP and 43218 protein sequences (having 135 \textit{GO}-terms) for MF respectively. Proposed \textit{ProtVecGen-Plus} based framework achieved an average \textit{F1}-score of 54.65 $\pm$ 0.15 and 65.91 $\pm$ 0.10  for BP and MF respectively. This is a significant improvement over existing state-of-the-art MLDA features \cite{MLDA_protFun} based model with corresponding average \textit{F1}-score of 51.66 $\pm$ 0.09 and 62.31 $\pm$ 0.09 respectively. However, the hybrid model outperformed all with an average \textit{F1}-score of 56.68 $\pm$ 0.13 and 67.12 $\pm$ 0.10 for BP and MF respectively.\vspace{2mm}

\item \textbf{Consistent results w.r.t. sequence lengths}: Proposed approach produces consistent results for a wide range of protein sequence lengths. 
\end{itemize}\vspace{1mm}

The rest of the paper is organized as follows. Section \ref{Methods} discusses the proposed framework architecture. Section \ref{hybrid_model} describes the proposed weighted hybrid model. This follows results  and discussion in Sec. \ref{res_discussion}. Finally, Sec. \ref{conclusion} concludes the paper.

\section{Proposed Method}\label{Methods}
Recent success of deep learning techniques such as Recurrent Neural Network (RNN) for the analysis of sequence and time series data has made a strong case for the use of these techniques for the analysis of protein sequences, such as protein structure prediction \cite{structure_lstm}, protein remote homology detection \cite{remote_homolgy, ProtDet-CCH}, and protein function prediction \cite{ProLanGo}. Motivated by this, the proposed framework uses RNN that processes the small segments of a protein sequence to model the protein for enhanced functional annotation. The proposed framework has the following components: 1) \textit{ProtVecGen} : a novel approach to protein vector construction based on small segments of protein sequences and bi-directional long short-term memory network (a class of RNN networks), and 2) \textit{Classification Model} : for predicting function(s) of protein sequences. These are described next. 

\subsection{Notations}
Let $\{GO_1, GO_2,...,GO_K\}$ denote the set of ``$K$" $GO$-terms (either for biological process or molecular function). Let $ P = \{(P_1, Y_1), (P_2,Y_2), ..., (P_n,Y_n)\} $ denote a database of $n$ labeled protein sequences, where $P_i$ denotes a protein sequence and $Y_i$ = $\{y_{i1}, y_{i2},...,y_{iK}\}$ denotes one-hot vector encoding representing a set of $GO$-term annotations corresponding to protein $P_i$; $y_{ik} = 1$ if $P_i$ exhibits $GO_{k}$ else 0.



\subsection{Protein Vector Construction}
For each amino acid sequence, a fixed-length feature vector is modeled by averaging all the possible segment vectors corresponding to the segments of a protein sequence. Protein feature vector construction involves three sub-steps: (i) Protein sequence segmentation, (ii) Segment vector generator and (iii) Protein vector generator. These are described next. 

\vspace{4mm}
\subsubsection{\textbf{Protein Sequence Segmentation}}\label{seq_seg}
Most existing approaches to protein function prediction such as  ProLanGo \cite{ProLanGo}, DeepGo \cite{DeepGo} and MLDA \cite{MLDA_protFun}, rely upon complete protein sequences for predicting functions. Nevertheless, the functionality of a protein is the consequence of a relatively small functional subsequence present within the protein sequence \cite{domain}; this sub-sequence can be considered as a latent pattern. More than often, a family of proteins having a common function share one or more latent patterns associated with the common function \cite{domain}.

Such a latent pattern is called \textit{conserved} if it can be directly associated with some \textit{GO}-term and is common in the family of proteins associated with the \textit{GO}-term. The part of a protein sequence that is not \textit{conserved} is called \textit{non-conserved} region. Since, the \textit{non-conserved} regions are assumed to be not associated with the functionality of a protein, they are considered as \textit{noise} and may adversely affect the vector formulation. The adverse affect is prominent especially in long sequences because, here, the noisy regions significantly dominate the conserved ones. In the proposed framework, we have incorporated segmentation approach to ensure efficient formulation of a protein vector. The aim of segmentation is to reduce the dominance of \textit{non-conserved} regions over relatively small \textit{conserved} regions in a protein sequence.

\begin{figure}[!t]
	\centering
	\includegraphics[width=3.5in]{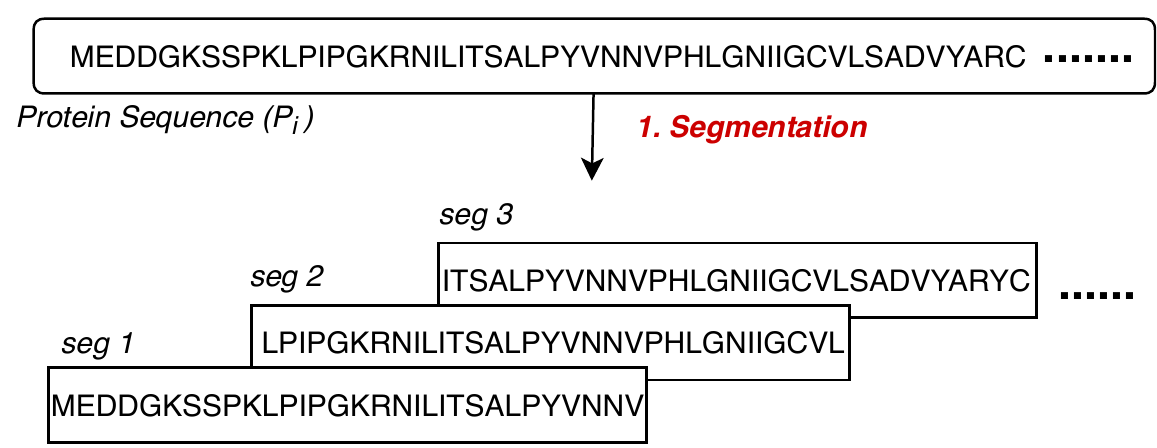}
	\caption{Protein Sequence Segmentation}
	\label{ProSegmen}
\end{figure}

Motivated by the above,  instead of using the complete full length amino acid sequences, an efficient segmented sequence approach is proposed for developing a global feature vector for a protein. As shown in Fig. \ref{ProSegmen}, each protein sequence $P_i$ having label set $Y_i$ is partitioned into equal sized segments with overlapping regions as $\phi_i = \{p_{i,1}, p_{i,2},...,p_{i,j},...\}$, where $\phi_i$ is the set of segments corresponding to protein $P_i$ and $p_{i,j}$ represents $j^{th}$ segment of protein $P_i$. The gap between two adjacent segments $p_{i,j}$ and $p_{i,j+1}$ is taken as 30 amino acids. Each segment $p_{i,j}$ is assigned the same label set $Y_i$ as the protein sequence. This constitutes the new training dataset of protein segments as $(p_{i,j}, Y_i)$, where $i \in \{1,2,..,n\}$ and $j \in \{1,2,...\}$. The choice of appropriate segment size has been discussed in details in Sec. \ref{App_seg_size}.

\vspace{4mm}

\subsubsection{\textbf{Protein Segment Vector Generator}}\label{sec_model}
This section describes the method to generate segment vectors based on Bi-Directional LSTM. A brief introduction to Bi-Directional LSTM is given next.
\vspace{4mm}

\textbf{Bi-Directional LSTM.}
Bi-Directional Long Short-Term Memory (Bi-LSTM) \cite{BiLstm} is a class of RNN, capable of learning long distance temporal dependencies. These are long chain of repeating units called memory cells. A small buffer element called cell state $C_t$ is the key to the network and passes through each memory cell in the chain. As the cell state passes through each memory cell, its content can be modified using special logics namely, forget logic and input logic. This modification depends upon $X$, which is a concatenation of the output of previous cell $h_{t-1}$ and the current input $x_t$:
\begin{equation}
X = concatenation[h_{t-1},x_t]\\
\end{equation}
A forget logic is used to remove the irrelevant information from the cell state as new input $x_t$ is encountered. It is composed of a single neural network layer with sigmoid activation function $\sigma$, which acts as a filter and produces a value in range [0,1] for each element in a cell state; the value represents the component of each element (i.e., information) to let through. Mathematically, it is described as:
\begin{equation}\label{forget_logic}
	f_t = \sigma(W_fX + b_f)
\end{equation}
where $W_f$ denotes weight matrix and $b_f$ denotes the bias vector with subscript term \textit{f} indicating forget gate. 

The input logic adds new information to the cell state. It is composed of two independent neural network layers with $\sigma$ and $tanh$ activations respectively. $\sigma$ does filtering and $tanh$ creates a vector of new elements. Mathematically, it is described as:
\begin{equation}\label{input_logic}
\begin{aligned}
i_t = \sigma(W_iX + b_i)\\
\bar{C_t} = tanh(W_cX + b_c)
\end{aligned}
\end{equation}
where,\\
$W_i$ and $W_c$ denotes weight matrices,\\
$b_i$ and $b_c$ denote bias vectors.

Next, the following equation is applied to update the information of a cell state by removing the information filtered out using forget logic and adding new information using input logic:
\begin{equation}\label{cell_state_update}
	C_t = (f_t \odot C_{t-1}) + (i_t \odot \bar{C_t})
\end{equation}
where, $\odot$ denotes element-wise multiplication. 

Finally, the hidden state $h_t$ at each memory cell is decided  based on the updated cell state $C_t$ and the output logic $o_t$, where $o_t$ is obtained using single neural network layer. They are described as:
\begin{equation}
	o_t = \sigma(W_oX + b_o)
\end{equation}
\begin{equation}
	h_t = o_t \odot tanh(C_t)
\end{equation}
where $W_o$ and $b_o$ denotes weight matrix and bias vector respectively with subscript term \textit{o} indicating output gate. In contrast to LSTM network, Bi-LSTM reads a sequence in both forward and backward directions. Like LSTM, it also generates a fixed-length feature vector for a sequence.
 \vspace{4mm}

\begin{figure}[!t]
	\centering
	\includegraphics[width=3.1in]{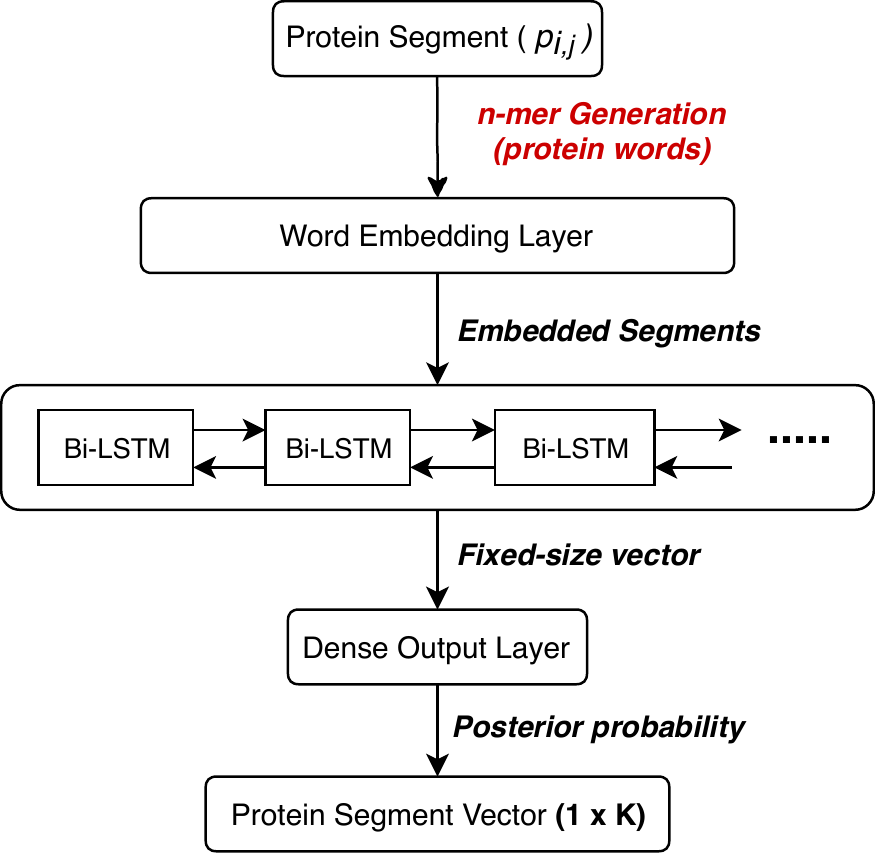}
	\caption{ProtSVG : Protein Segment Vector Generator}
	\label{SegVecGenerator}
\end{figure}

\textbf{Protein Segment Vector Generator.} 
We propose a method Protein Segment Vector Generator (ProtSVG) to convert protein segments to a segment vector. In ProtSVG, each segment $p_{i,j} \in \phi_i$ of protein sequence $P_i$ is modeled independently irrespective of its neighboring segments and is represented as protein segment vector $\textbf{p}_{i,j}$. Encoded vector representation for a protein segment is formulated based on bi-directional LSTM as discussed above. Figure \ref{SegVecGenerator} describes ProtSVG model. It consists of three layers: (i) embedding layer as input layer, (ii) bi-directional LSTM layer,  and (iii) dense layer with sigmoid activation function. The model is trained using protein segment pairs $(p_{i,j}, Y_i)$ from the training dataset.

Each protein segment is decomposed into $n$-mers (with $n = 4$) to formulate protein words. These words are placed in the same order in which they are found in the protein segment to form a sequence. These sequences of words are fed as input to the embedding layer which outputs dense representation for each protein segment. The embedding layer generates a matrix of size ($l$ - 4 + 1) x 32, where $l$ denotes the fixed length of segments and 32 denotes the size of embedding. Padding is done on the last segment of a protein sequence if its length $<l$. Next, the bi-directional LSTM layer has 70 memory cells in one block, which receive dense representations from the embedding layer. Dropout \cite{dropout} was applied in bi-directional LSTM layer and the proportion of disconnection was 0.3. This network was implemented  using Keras 2.0.6\footnote{\url{https://github.com/keras-team/keras}} with the backend of TensorFlow.

The trained ProtSVG model is then used to produce a fixed-size vector $\textbf{p}_{i,j} = \{o_1, o_2,...,o_K\}$ for each protein segment  $p_{i,j}$. Here, $K$ is the total number of $GO$-terms for either molecular function or biological process, and 
$o_k$ denotes the posterior probability of $GO_k$ given the protein segment $p_{i,j}$:
\begin{equation}
\begin{aligned}
o_k(p_{i,j}) = Pr(GO_k  \arrowvert  p_{i,j})
\end{aligned}
\end{equation}

\vspace{4mm}
\subsubsection{\textbf{Feature Vector Construction}}
To generate the global feature vector of an entire protein sequence,  we average its segment vectors as obtained from ProtSVG. The entire process is described in Algorithm \ref{vector_protein}. We call this algorithm, ProtVecGen. In step 1, the entire protein sequence is segmented into equal size segments of size $s$. Segments $\{p_{i,1}, p_{i,2},...,p_{i,j},...\}$ created for each amino acid sequence $P_i$ in the training set are then fed to the trained ProtSVG model. This yields segment vector $\textbf{p}_{i,j}$ corresponding to segment $p_{i,j}$ as shown in step [2-4]. In step 5, all such $\textbf{p}_{i,j}$s corresponding to protein $P_i$ are averaged to  yield the global feature vector $\textbf{P}_{i,s} = \{O_1, O_2,...,O_{K}\}$, where:
\begin{equation}
\begin{aligned}
O_k = \frac{1}{\arrowvert \phi_i \arrowvert}\sum_{p_{i,j} \in \phi_i}o_k(p_{i,j})\\
j=\{1,2,...\}, \hspace{3mm}  k=\{1,2,...,K\}
\end{aligned}
\end{equation}
here, $\arrowvert \phi_i\arrowvert$ indicates the number of segments for protein $P_i$ and $O_k$ is the mean posterior probability for $GO_k$. This process  allows us to generate equal-sized descriptor for each protein sequence  irrespective of its length.

The benefits of the proposed approach are three-fold;

1) It retains relative ordering of $n$-mers in the protein sequence, which is not the case with frequently used \textit{tf-idf} weight scheme. 

2) It fragments the protein sequence into small segments, transforming each such segment into a fixed size feature vector. This enables machine learning models to learn the latent patterns conserved within small regions without getting affected by remaining \textit{non-conserved} parts of the complete sequence. Thus, more a machine learning models sees such a  recurrent \textit{conserved} pattern, the easier it becomes for the model to associate the \textit{conserved} pattern with the specific protein functionality.  

3) The segmentation approach also avoid the adverse effects due to high amounts of padding for short protein sequences and truncation of long sequences as in \cite{remote_homolgy,ProtDet-CCH}. The latter may result in information loss or even dropping of some \textit{conserved} pattern.

\begin{algorithm}[t]
	\caption{ProtVecGen: Algorithm for Protein Vector Generator.}
	\begin{algorithmic}[1]\label{vector_protein}
		\renewcommand{\algorithmicrequire}{\textbf{Input:}}
		\renewcommand{\algorithmicensure}{\textbf{Output:}}
		\REQUIRE Protein Sequence $P_i$ and Segment size \textit{s}
		\ENSURE  Protein Sequence Vector  $\textbf{P}_{i,s}$
		\STATE Partition $P_i$ into set of fixed sized segments  $\phi_i = \{p_{i,1}, p_{i,2},...,p_{i,j},...\}$.
		\FOR {{$p_{i,j}$ in $\phi_i$}}
		\STATE Generate segment vector $\textbf{p}_{i,j}$ using protein segment vector generator ProtSVG.
		\ENDFOR
		\vspace{1mm}
		\STATE $\textbf{P}_{i,s} \leftarrow \frac{1}{|\phi_i|}\sum\limits_{p_{i,j} \in \phi_i}\textbf{p}_{i,j}$
		\vspace{1mm}
		\RETURN $\textbf{P}_{i,s}$ 
	\end{algorithmic} 
\end{algorithm}

\subsection{Classification based on Multi-sized Segment Feature Vectors}\label{Query_model}
Because the size of the \textit{conserved} patterns can vary from one sequence to another, partitioning a sequence based on a fixed segment size may split a \textit{conserved} region across two segments. In order to prevent this, we partitioned protein sequences based on multiple segment sizes so that each conserved pattern is preserved in at least one of the segments. Protein vectors were generated separately for each segment size. For a given sequence, the protein vectors corresponding to its different-sized segments are then concatenated to produce a merged vector, which is finally used for training.  

We call this method ProtVecGen-Plus as described in Algorithm \ref{vector_fusion}.  In steps [2-4], for each protein sequence $P_i$, separate protein vectors for three segment sizes 100, 120, and 140 are generated as $\textbf{P}_{i,100}$, $\textbf{P}_{i,120}$, and $\textbf{P}_{i,140}$ respectively using Algorithm \ref{vector_protein}. These are then concatenated in step 5 ($||$ denotes concatenation operator) to yield vector $\textbf{P}_i^{+}$, which is finally used for training. The  dimension of $\textbf{P}_i^{+}$ is $3K$, where $K$ denotes the total number of $GO$-terms as defined earlier. The choice of ${100, 120, 140}$ as segment sizes was made after evaluating a wide range of segment sizes from 60 to 700; detailed discussion regarding this is given in Sec. \ref{App_seg_size}.

\begin{figure}[!t]
	\centering
	\includegraphics[width=2.3in]{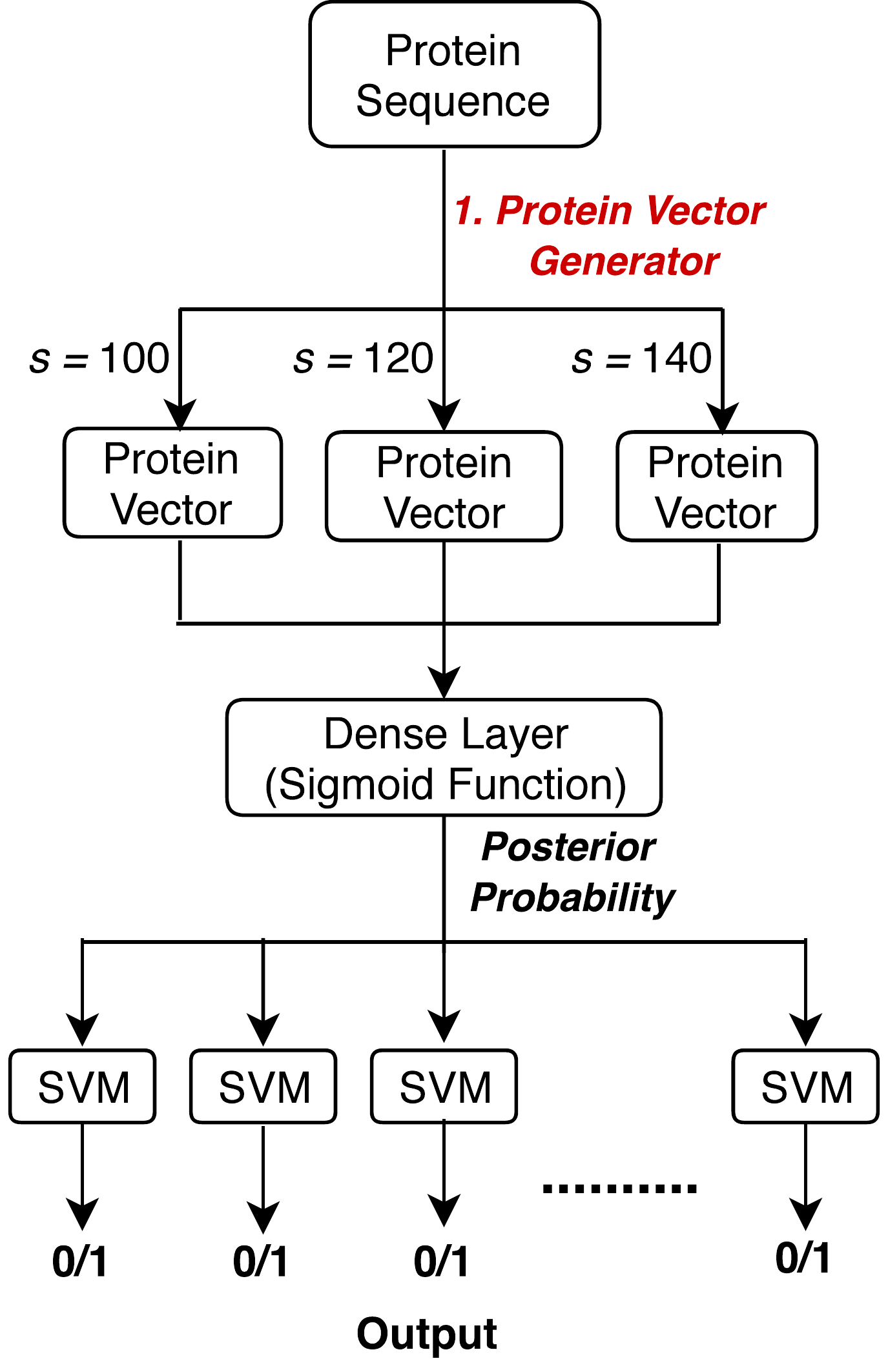}
	\caption{Proposed Framework}
	\label{ProposedModel}
\end{figure}

The proposed framework is shown in Fig. \ref{ProposedModel}. It consists of three layers (i) Protein vector generator layer, (ii) a fully-connected neural network layer and (iii) an SVM layer, connected sequentially as shown. The first layer generates protein vectors as described in Algorithm \ref{vector_fusion}. The generated protein vectors  $\textbf{P}_i^{+}$s are fed to the fully-connected neural network layer. The neural network layer consists of a single hidden layer followed by a dense output layer with sigmoid activation function. Binary cross-entropy and adam are the loss function and optimizer respectively. The output of the neural network layer is fed to the SVM layer. However, the neural network and SVM layers train separately.

\begin{algorithm}[t]
	\caption{ProtVecGen-Plus: Algorithm for construction of protein vector based on feature concatenation.}
	\begin{algorithmic}[1]\label{vector_fusion}
		\renewcommand{\algorithmicrequire}{\textbf{Input:}}
		\renewcommand{\algorithmicensure}{\textbf{Output:}}
		\REQUIRE Protein Sequence as $P_i$
		\ENSURE  Protein Sequence Vector as $\textbf{P}_i^{+}$
		\STATE \textit{Initialisation} : 
		\\$segSize$ = \{100, 120, 140\}
		\FOR {\textit{size} in \textit{segSize}}
		\STATE $\textbf{P}_{i,size}$ = ProtVecGen($P_i, size$)
		\ENDFOR
		\STATE $\textbf{P}_i^{+} = [\textbf{P}_{i,100} || \textbf{P}_{i,120} || \textbf{P}_{i,140}$]
		\RETURN $\textbf{P}_i^{+}$ 
	\end{algorithmic} 
\end{algorithm} 

First, the neural network trains on $\textbf{P}_i^{+}$s to produce a posterior probability $Pr(GO_k|\textbf{P}_i^{+})$ for each individual functional class $GO_k$ given $\textbf{P}_i^{+}$, where $k = \{1,2,...,K\}$. Next, the SVM layer is trained. Here, a separate SVM\footnote{Python library scikit-learn\cite{scikit} is used to realize SVMs. Default SVM setup using RBF kernel with hinge loss as loss function and regularization parameter $C = 1.0$ is chosen for the purpose of training.} is trained for each $GO$ term $GO_k$  that produces a binary output, where value 1  predicts that $\textbf{P}_i^+$ is annotated with $GO_k$ and value 0 predicts otherwise. To train a separate SVM for each $GO_k$, where $k = \{1,2,...,K\}$, only the output (i.e., posterior probabilities $Pr(GO_k|\textbf{P}_{i}^{+})$, where $i \in {1,2,...,n}$) generated for $GO_k$ from the above trained neural network layer is used as an input to the SVM layer. Using SVM eliminates the need to manually identify a threshold for converting continuous probabilities to discrete binary values.

\section{Weighted Hybrid Framework Approach}\label{hybrid_model}
While ProtVecGen produces better results than state-of-the-art MLDA feature \cite{MLDA_protFun} based classifier, and ProtVecGen-Plus does even better (detailed results are discussed in Sec. \ref{res_discussion}), a new hybrid framework is proposed to further boost the predictive performance. The proposed hybrid framework combines the proposed ProtVecGen-Plus based model with one based on MLDA (Multi-label Linear Discriminant Analysis) features \cite{MLDA_protFun}. The MLDA based model is trained using MLDA features on a two-layer neural network classifier as described before in the proposed model. There are two reasons of using MLDA based model: (i) Reference \cite{MLDA_protFun} is the state-of-the-art for multi-label protein function prediction using machine learning techniques, and (ii) it produces good results for classifying short and mid-range protein sequences as elaborated in Sec. \ref{Seq_len_effect}.  The complete hybrid framework is shown in Fig. \ref{HybridModel}. We first describe MLDA in Sec. \ref{MLDA}. Then, the proposed weighted scheme to combine the predictions from the two models has been  described in Sec. \ref{Hybrid_Framework}. 

\subsection{Multi-Label Linear Discriminant Analysis}\label{MLDA}
Multi-Label Linear Discriminant Analysis (MLDA) \cite{MLDA} is reduction technique and a generalization of the classical Linear Discriminant Analysis (LDA) method for multi-label scenario. Specifically, classical LDA assumes that each data element belongs to one class only, whereas MLDA also takes care of an element belonging to multiple classes. Contribution of such data elements towards each class is taken separately. Like LDA, MLDA also projects data from feature space $F$ to a subspace $F_s$:
\begin{equation}
	\textbf{y} = U^T\textbf{x}
\end{equation}
where, $\textbf{x}$ is a data element in a high dimension feature space $F$, $\textbf{y}$ is the projected element in low dimension subspace $F_s$, and $U^T$ is the projection matrix. 

Like LDA, in MLDA the projection matrix $U^T$ is obtained by solving the eigenvalue problem for matrix:
 \begin{equation}
 S_w^{-1}S_b
 \end{equation}
where, $S_b$ and $S_w$ denote  between-class and within-class scatter matrices respectively. 
In case of MLDA, these matrices are calculated as:

\begin{equation}\label{scatter_mat1}
\begin{aligned}
S_b = \sum_{k=1}^{K}\Bigg(\sum_{i=1}^{n}y_{ik}\Bigg)(\bar{\textbf{m}}_k - \bar{\textbf{m}})(\bar{\textbf{m}}_k - \bar{\textbf{m}})^T\\
S_w = \sum_{k=1}^{K}\Bigg(\sum_{i=1}^{n}y_{ik}(\textbf{x}_i - \bar{\textbf{m}}_k)(\textbf{x}_i - \bar{\textbf{m}}_k)^T\Bigg)
\end{aligned}
\end{equation}
where, $K$ is the cardinality of $GO$-term and $n$ is the number of sequences. $y_{ik} = 1$, if $\textbf{x}_i$ has $GO_k$ as its annotation and 0 otherwise.

$\bar{\textbf{m}}_k$ and $\bar{\textbf{m}}$ denote class mean of $GO_k$ and global mean respectively. They are calculate as;
\begin{equation}
\bar{\textbf{m}_k} = \frac{\sum_{i=1}^{n}y_{ik}\textbf{x}_i}{\sum_{i=1}^{n}y_{ik}}
\quad \text{,} \quad
\bar{\textbf{m}} = \frac{\sum_{k=1}^{K}\sum_{i=1}^{n}y_{ik}\textbf{x}_i}{\sum_{k=1}^{K}\sum_{i=1}^{n}y_{ik}}
\end{equation}

\subsection{Function Prediction based on Hybrid Framework}\label{Hybrid_Framework}
\begin{figure}[!t]
	\centering
	\includegraphics[width=2.7in]{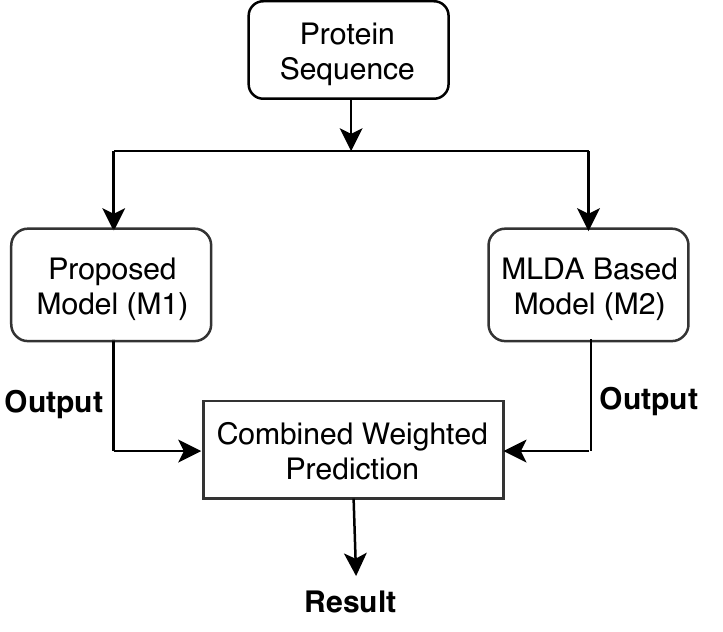}
	\caption{Hybrid Framework}
	\label{HybridModel}
\end{figure}

Functional annotations of proteins is done using the combined weighted predictions of: 1) The proposed model (represented as M1) and 2) MLDA based model \cite{MLDA_protFun} (represented as M2) as shown in Fig. \ref{HybridModel}. The MLDA-based model uses \textit{tf-idf} features for classification  \cite{MLDA_protFun}. Each protein sequence is decomposed into \textit{n}-mers, based on which the corresponding \textit{tf-idf} weights are computed. Each weight highlights the importance of an \textit{n}-mer (\textit{n} = 3) in a protein sequence with respect to the entire input dataset. Because, the \textit{tf-idf} technique produces a sparse and high-dimensional feature vector, hence feature reduction is needed. Application of MLDA on \textit{tf-idf} features produces a dense and low dimensional feature vector with size of vector equal to ($K$-1), which is eventually used for classification. The combined weighted prediction is computed as:
\begin{equation}\label{weight_formula}
	z_{k} = \alpha(z_{k}^1) + (1 - \alpha)(z_{k}^2), \hspace{6mm} \alpha \in [0,1]
\end{equation}
where, $z_{k}^1$ and $z_{k}^2$ denote the predictions by M1 and M2 respectively for $k^{\text{th}}$ $GO$-term, and $\alpha$ denotes the trade-off parameter. It is formulated empirically using avg. $F1$-scores of M1 and M2:
\vspace{1mm}
\begin{equation}\label{alpha}
	\alpha = \frac{\text{avg. }F1\text{-}score\text{(M1)}}{\text{avg. }F1\text{-}score\text{(M1)} + \text{avg. }F1\text{-}score\text{(M2)}}
\end{equation}\vspace{0.4mm}
where, avg. $ F1 $-score(M1) and avg. $ F1 $-score(M2) denote the average F1-scores of M1 and M2 respectively.

\section{Results and Discussion}\label{res_discussion}
We have evaluated the proposed models using protein sequences from UniProtKB database \cite{uniprot}, which consists around 558125 protein sequences reviewed and annotated with $GO$-terms. Out of these, we randomly chose 103683 protein sequences labeled with BP and 91690 protein sequences labeled with MF for experiments. Further, we removed those $GO$ terms which annotated less than 200 sequences. Protein sequences not annotated with at least one of the remaining $GO$ terms were dropped. This left us with 58310 sequences (with 295 \textit{GO}-terms) for BP prediction and 43218 sequences (with 135 \textit{GO}-terms) for MF prediction.\footnote{The dataset have been made available at \url{http://bit.ly/2O2NMJ0}.} For experiments, dataset was split into test dataset (25\%) and train dataset (75\%). 
\subsection{Evaluation Methods and Metrices}
Metrics \textit{precision}, \textit{recall}, and \textit{F1-score} were used to evaluate  the performance of the proposed models \cite{Metrices}. Though these are well known, they are also being defined herewith to emphasize their use in the case of multi-label classification. Let $Y_i = \{y_{i1}, y_{i2},...\}$ be the actual set of $GO$-terms annotating protein $P_i$. Let $Z_i = \{z_{i1}, z_{i2},...\}$ be the corresponding  predicted set of $GO$-terms.
\begin{enumerate}
	\item Average Precision: \textit{Precision} indicates the fraction of predicted $GO$-terms in set $Z_i$ that are correct. Average precision is the mean of precision values for all the data samples.
	\begin{equation}
	\text{\textit{Average Precision}} = \frac{1}{n}\sum_{i=1}^{n}\frac{\arrowvert Y_i \bigcap Z_i \arrowvert}{\arrowvert Z_i\arrowvert}
	\end{equation}
	\item  Average Recall: \textit{Recall} captures the fraction of $GO$-terms in actual set $Y_i$ that are predicted. Average recall is the mean of recall values for all the data samples.
	\begin{equation}
	\text{\textit{Average Recall}} = \frac{1}{n}\sum_{i=1}^{n}\frac{\arrowvert Y_i \bigcap Z_i \arrowvert}{\arrowvert Y_i\arrowvert}
	\end{equation}
	\item Average F1-Score: F1-Score is the harmonic mean of precision and recall values. Average f1-score is the mean of f1-score values for all the data samples.
	\begin{equation}
	\text{\textit{Average F1-score}} = \frac{1}{n}\sum_{i=1}^{n}\frac{2\arrowvert Y_i \bigcap Z_i \arrowvert}{\arrowvert Y_i\arrowvert + \arrowvert Z_i\arrowvert}
	\end{equation}
\end{enumerate}
For computing the above metrics, the values were averaged over 10 runs. 

\subsection{Choosing Appropriate \textit{n}-Mer}
The choice of \textit{n}-mer has a direct effect on the performance of the proposed framework. We experimentally evaluated  \textit{n} = \{3, 4, 5\} sized \textit{n}-mers (see Table \ref{n_mer_choice}) and found size \textit{n} = 4 to be the best in terms of prediction performance. The results shown in Table \ref{n_mer_choice} were obtained on segment size of 120. Similar performance behaviour is exhibited for both biological process and molecular function.
\begin{table*}[!t]
	\caption{Classification Accuracy Based on Different \textit{n}-mers.}
	\label{n_mer_choice}
	\centering
	\begin{tabular}{|p{0.81cm}|p{0.8cm}|p{1.91cm}|p{2.27cm}|p{2.19cm}|p{1.91cm}|p{2.27cm}|p{2.19cm}|}\hline 
		& \multicolumn{1}{c|}{} & \multicolumn{3}{c|}{Biological Process} & \multicolumn{3}{c|}{Molecular Function}\Tstrut\\ [0.4ex]
		\hline 
		S. No. & \textit{n}-mer & Avg. Recall (\%) & Avg. Precision (\%) & Avg. F1-Score (\%) & Avg. Recall (\%) & Avg. Precision (\%) & Avg. F1-Score (\%)\Tstrut\\
		\hline
		1 & 3-mer & 28.22 $\pm$ 0.55 & 31.43 $\pm$ 0.58 & 28.86 $\pm$ 0.59  & 38.83 $\pm$ 0.67 & 41.81 $\pm$ 0.67 & 39.53 $\pm$ 0.63 \Tstrut\\ [0.4ex]
		\hline
		2 & 4-mer & \textbf{53.58 $\pm$ 0.11} & \textbf{55.07 $\pm$ 0.07} & \textbf{52.64 $\pm$ 0.08} & \textbf{64.06 $\pm$ 0.21} & \textbf{64.81 $\pm$ 0.06} & \textbf{63.18 $\pm$ 0.11} \Tstrut\\ [0.2ex]
		\hline
		3 & 5-mer & 51.81 $\pm$ 0.38 & 54.36 $\pm$ 0.35 & 51.39 $\pm$ 0.29 & 59.91 $\pm$ 0.38 & 61.25 $\pm$ 0.51 & 59.61 $\pm$ 0.37 \Tstrut\\ [0.4ex] 
		\hline
	\end{tabular}
\end{table*}

\begin{figure}[!t]
	\centering
	\subfloat[Biological Process]{\includegraphics[width=3.4in]{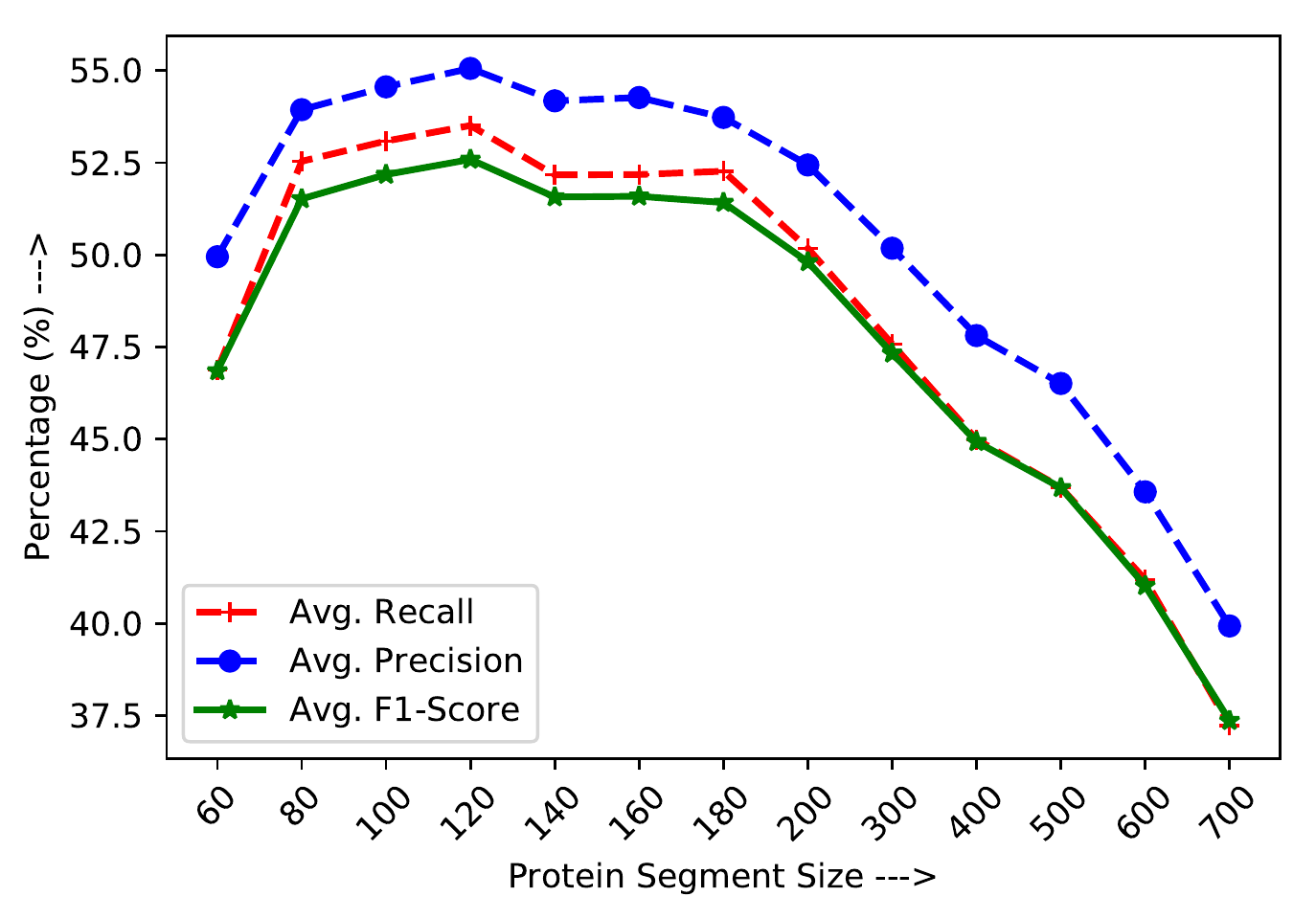}\label{seg_effect_BP}}
	
	\subfloat[Molecular Function]{\includegraphics[width=3.4in]{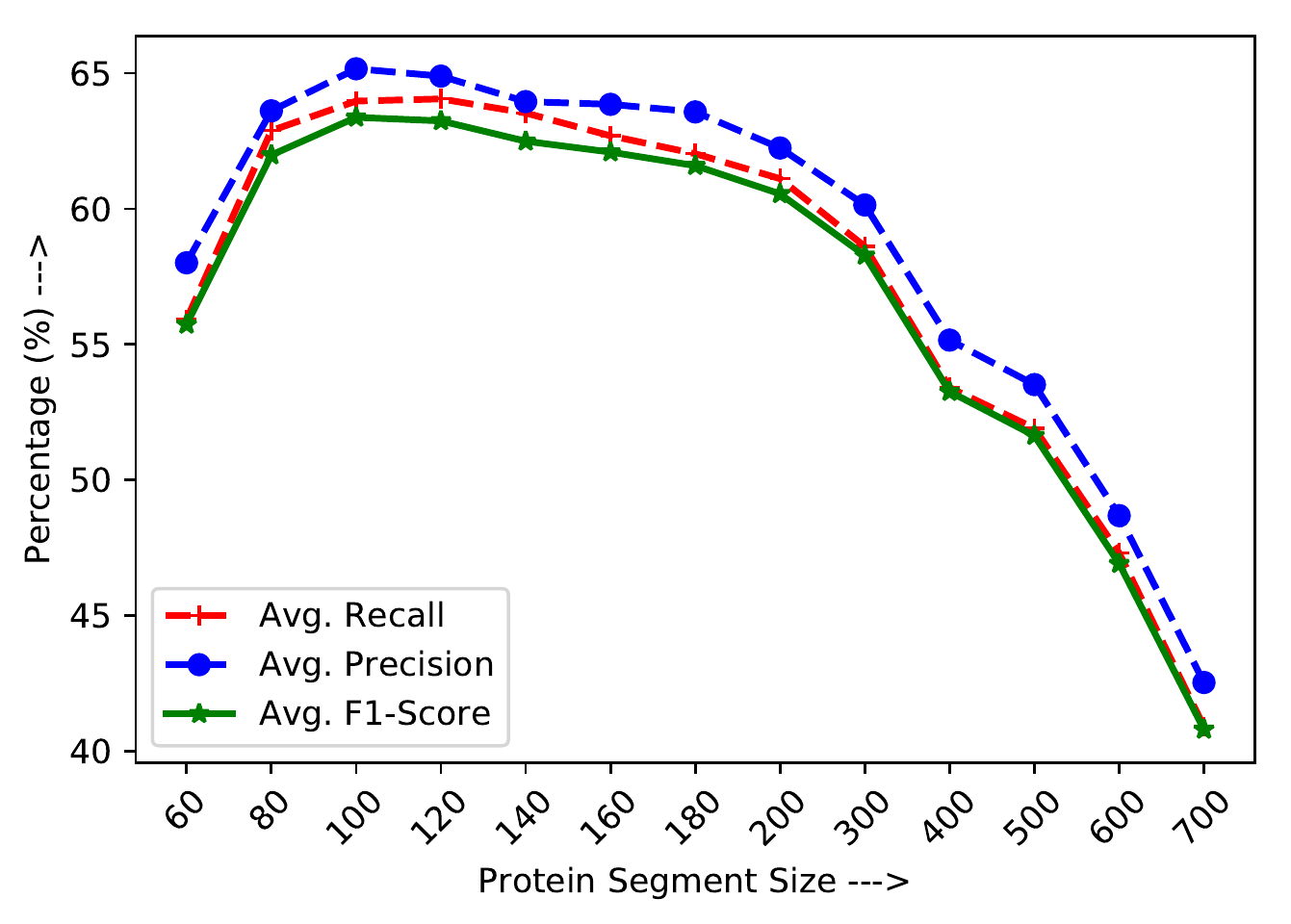}\label{seg_effect_MF}}
	
	\caption{Effect of segment size on the overall performance of the proposed framework for both biological process and molecular function.}
	\label{SegEffect}
\end{figure}

\subsection{Choosing Appropriate Segment Size} \label{App_seg_size}
The choice of appropriate segment size is a crucial task for the proposed approach. The best segment size was chosen based on empirical evaluation of the proposed framework with different segment sizes in the range [60-700]. Figure \ref{seg_effect_BP} shows values (in percentage) of average precision, average recall and average \textit{f1} score for different segment sizes for biological process prediction.  Figure \ref{seg_effect_MF} shows the corresponding values for molecular function prediction. Overall performance drops significantly with segment size above 180.

\begin{figure*}[!t]
	\centering
	\subfloat[Biological Process] {\includegraphics[width=2.95in]{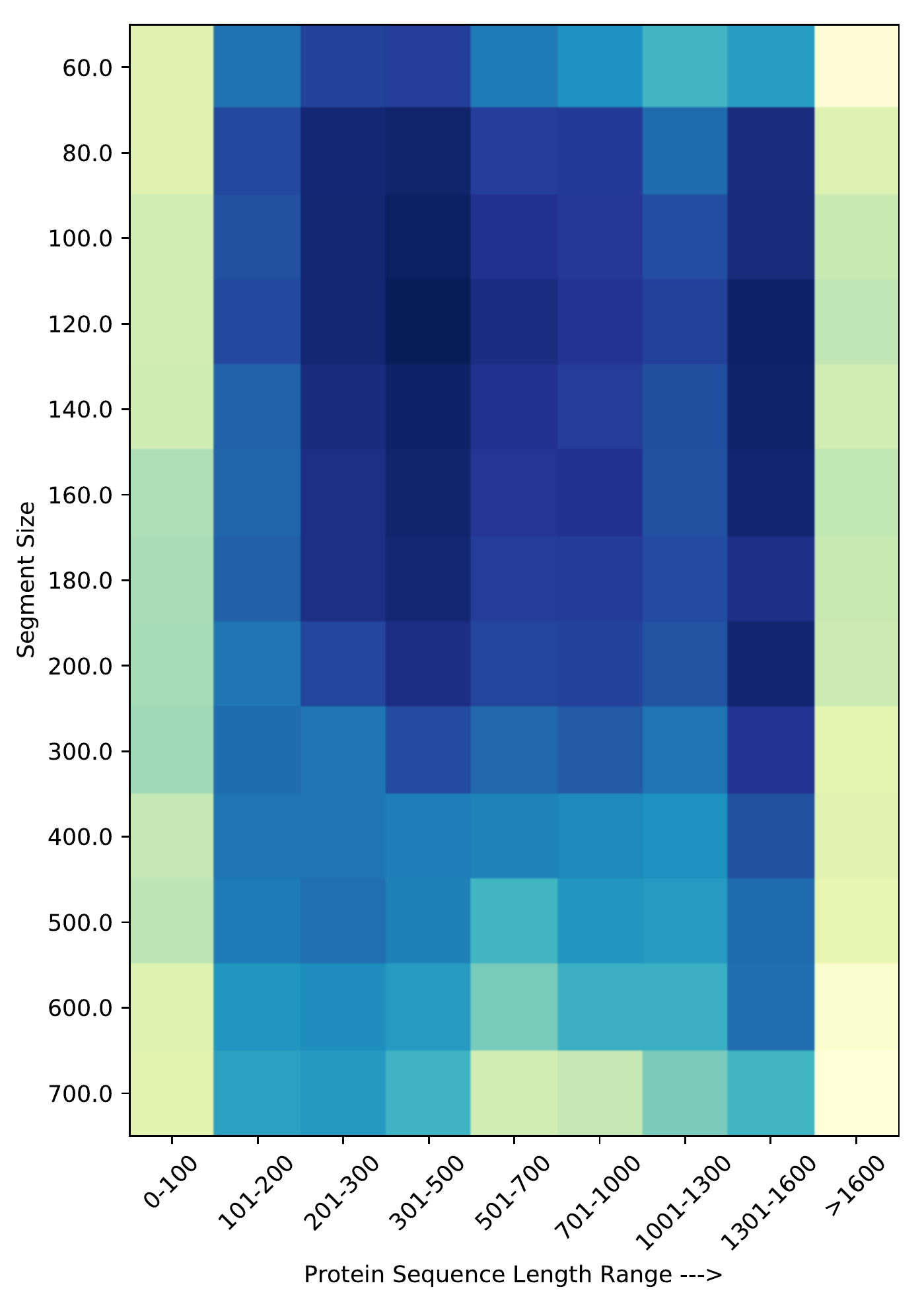}
		\label{heat_map_BP}}
	\hfil
	\subfloat[Molecular Function] {\includegraphics[width=2.95in]{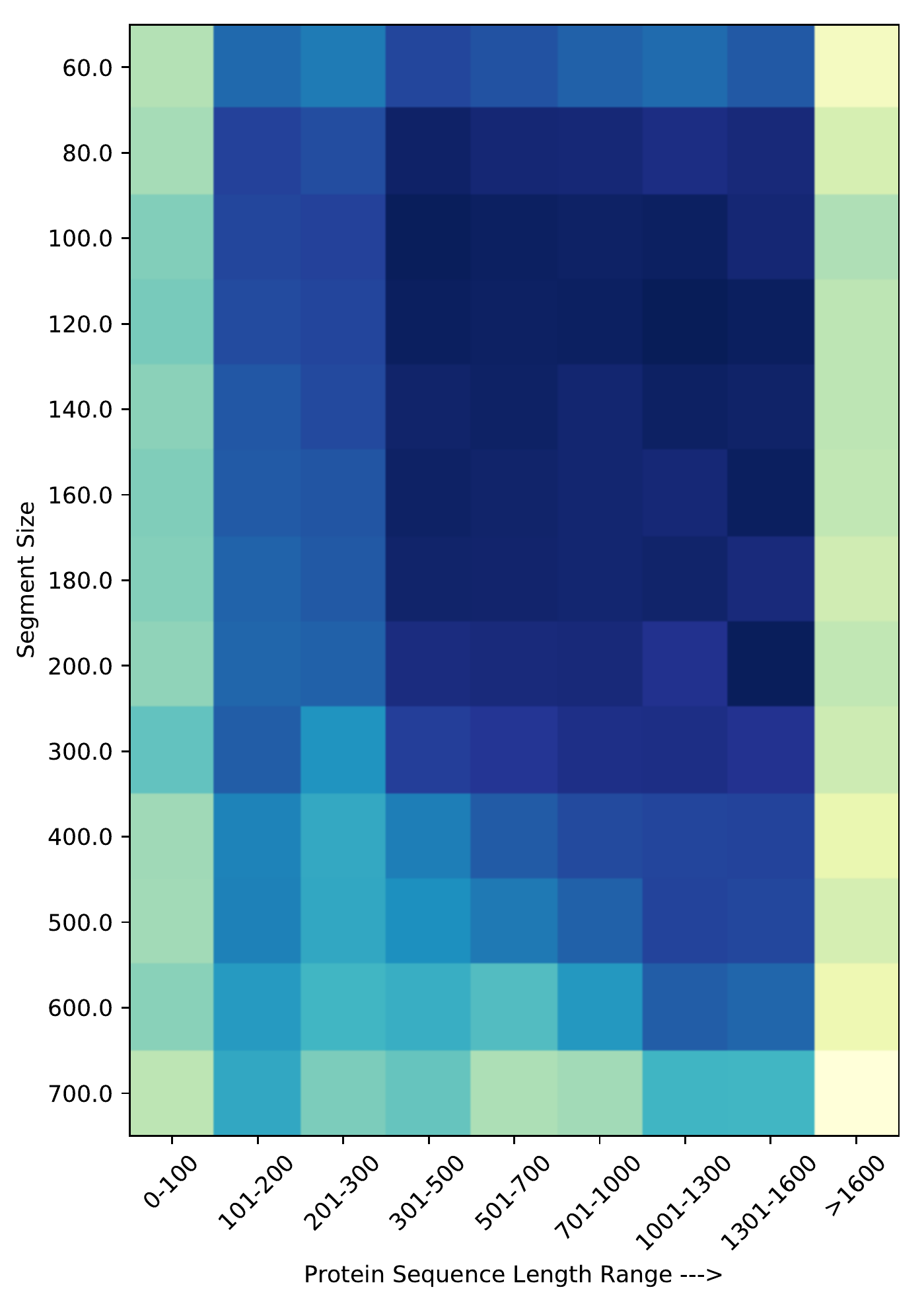}
		\label{heat_map_MF}}
	\caption{Performance of the proposed framework for protein sequences in different length ranges with respect to different segment sizes. Figure (\ref{heat_map_BP}) shows performance for biological process, while (\ref{heat_map_MF}) is for molecular function. Darker regions indicates high performance. Performance corresponding to protein sequences of short (0-100 amino acids) and long ($\textgreater$1600) sequence length are low compared to the rest. Performance of models \textit{ProtVecGen-100} (based on segment size 100) and \textit{ProtVecGen-120} (based on segment size 120) are close and superior to the others.}
	\label{heat_map}
\end{figure*}

As shown in Fig. \ref{SegEffect}, the proposed framework yields best results, for both BP and MF, in the segment size range of [80-180]. The same is conveyed by Fig. \ref{heat_map}, which presents a further detailed performance behaviour of the framework for sequence length range vs. segment size. Motivated by individual model performances with respect to different segment sizes, we chose sizes 100, 120 and 140 to segment each protein sequence. Further, vectors corresponding to the three segment sizes were then concatenated to formulate global protein vector based on multi-segment approach for all experiments, as discussed in Sec. \ref{Query_model}. 

\subsection{Single Segment vs Multi Segment Approach}\label{single vs multi}
We also compared the effect of the three chosen segment sizes individually. Performance of feature vector \textit{ProtVecGen}-100 (based on segment size 100) for the BP is shown in Table \ref{acc_bio} along row 4 under the heading Biological Process. Similarly, the performances of \textit{ProtVecGen}-120 (based on segment size 120)  and \textit{ProtVecGen}-140 (based on segment size 140) feature vectors are shown in rows 5  and 6  respectively. \textit{ProtVecGen-Plus} feature vector formulated based on multiple segments of size 100, 120, and 140 is also compared as shown in row 7 of Table \ref{acc_bio}. Results in the remaining rows are discussed later. Columns under heading Molecular Function in Table \ref{acc_bio} shows the corresponding values for MF.

\textit{ProtVecGen-Plus} feature based classifier yields overall average \textit{F1}-scores of 54.65 $\pm$ 0.15 and 65.91 $\pm$ 0.10 for BP and MF respectively. This is a significant improvement over the second best classifier trained using \textit{ProtVecGen}-120 for the BP with average \textit{F1}-scores of 52.64 $\pm$ 0.08. For the case of MF, \textit{ProtVecGen}-100 feature based classifier is the next best with average \textit{F1}-score of 63.39 $\pm$ 0.15. These results clearly demonstrate the superiority of multi-segment based \textit{ProtVecGen+Plus} feature over the other three based on single segment sizes.

\begin{table*}[!t]
	\caption{Classification Accuracy.}
	\label{acc_bio}
	\centering
	\begin{tabular}{|p{0.36cm}|p{3.37cm}|p{1.53cm}|p{1.78cm}|p{1.61cm}|p{1.53cm}|p{1.78cm}|p{1.61cm}|}\hline 
		& \multicolumn{1}{c|}{} & \multicolumn{3}{c|}{Biological Process} & \multicolumn{3}{c|}{Molecular Function}\Tstrut\\ [0.4ex]
		\hline  \Tstrut
		S. No. & Approach & Avg. Recall (\%) & Avg. Precision (\%) & Avg. F1-Score (\%) & Avg. Recall (\%) & Avg. Precision (\%) & Avg. F1-Score (\%)\\
		\hline \Tstrut
		1 & Bi-LSTM + NN (Complete) & 31.63 $\pm$ 0.11 & 33.34 $\pm$ 0.11 & 31.59 $\pm$ 0.12  & 37.81 $\pm$ 0.06 & 38.5 $\pm$ 0.05 & 37.43 $\pm$ 0.05 \\ [0.4ex]
		\hline \Tstrut
		2 & SGT \cite{SGT} + NN & 40.95 $\pm$ 0.13 & 40.91 $\pm$ 0.14 & 39.53 $\pm$ 0.12 & 52.82 $\pm$ 0.15 & 48.46 $\pm$ 0.14 & 48.69 $\pm$ 0.16 \\ [0.4ex]
		\hline \Tstrut
		3 & MLDA \cite{MLDA_protFun} + NN & 49.42 $\pm$ 0.10 & 52.61 $\pm$ 0.14 & 49.27 $\pm$ 0.12 & 58.29 $\pm$ 0.16 & 60.20 $\pm$ 0.20 & 57.91 $\pm$ 0.17 \\ [0.4ex] 
		\hline \Tstrut
		4 & ProtVecGen-100 + NN & 53.15 $\pm$ 0.09 & 54.42 $\pm$ 0.11 & 52.11 $\pm$ 0.10 & 63.93 $\pm$ 0.25 & 65.25 $\pm$ 0.11 & 63.39 $\pm$ 0.15 \\
		[0.4ex]
		\hline \Tstrut
		5 & ProtVecGen-120 + NN & 53.58 $\pm$ 0.11 & 55.07 $\pm$ 0.07 & 52.64 $\pm$ 0.08 & 64.06 $\pm$ 0.21 & 64.81 $\pm$ 0.06 & 63.18 $\pm$ 0.11 \\ [0.4ex]
		\hline \Tstrut
		6 & ProtVecGen-140 + NN & 52.34 $\pm$ 0.12 & 54.22 $\pm$ 0.06 & 51.66 $\pm$ 0.09 & 63.16 $\pm$ 0.21 & 63.93 $\pm$ 0.08 & 62.31 $\pm$ 0.09 \\ [0.4ex]
		\hline\Tstrut 
		7 & ProtVecGen-Plus + NN & \textbf{56.42 $\pm$ 0.24} & \textbf{56.65 $\pm$ 0.12} & \textbf{54.65 $\pm$ 0.15} & \textbf{66.93 $\pm$ 0.21} & \textbf{67.42 $\pm$ 0.10} & \textbf{65.91 $\pm$ 0.10} \\ [0.2ex]
		\hline \Tstrut
		8 & ProtVecGen-Plus + MLDA + NN & \textbf{58.19 $\pm$ 0.14} & \textbf{58.80 $\pm$ 0.13} & \textbf{56.68 $\pm$ 0.13} & \textbf{68.62 $\pm$ 0.12} & \textbf{68.27 $\pm$ 0.13} & \textbf{67.12 $\pm$ 0.10} \\
		\hline
	\end{tabular}
\end{table*}

\subsection{Effect of the length of protein sequences}\label{Seq_len_effect}
\begin{figure*}[!t]
	\centering
	\subfloat[Biological Process] {\includegraphics[width=3.5in]{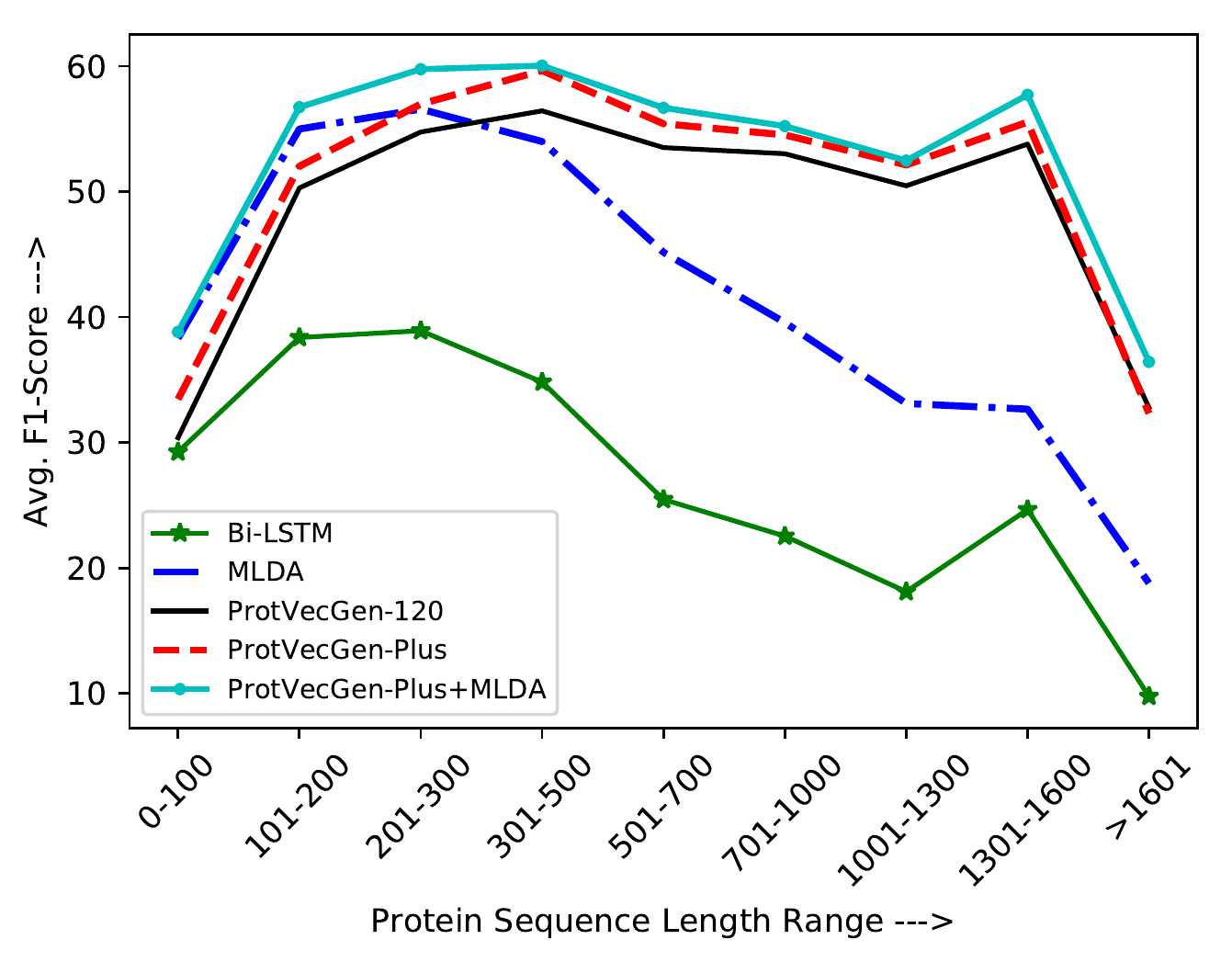}%
		\label{f1-score_BP}}
	\hfil
	\subfloat[Molecular Function]
	{\includegraphics[width=3.5in]{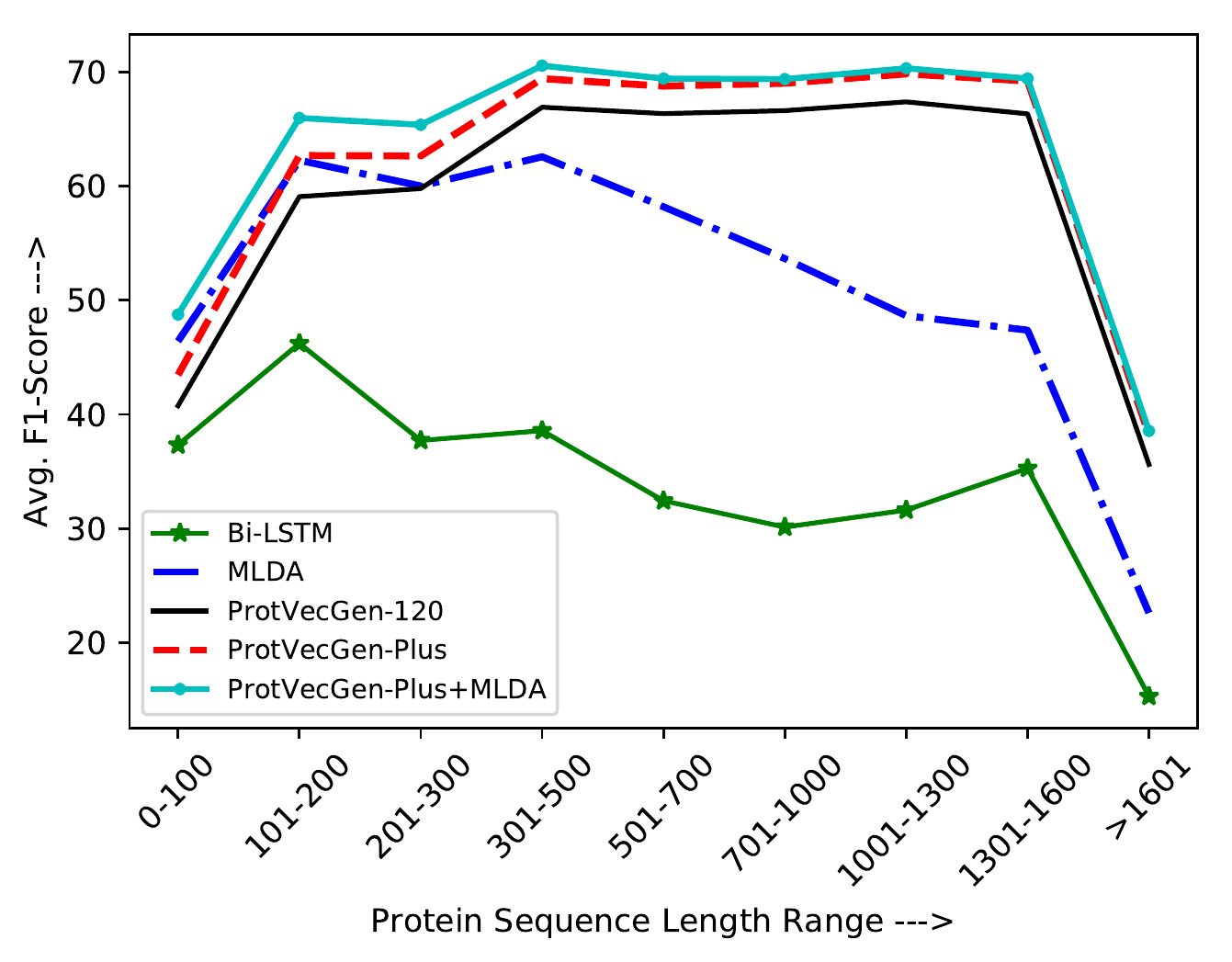}%
		\label{f1-score_MF}}
	\caption{Average \textit{F1}-Score for biological process and molecular function.}
	\label{F1-score}
\end{figure*}

\begin{figure*}[!t]
	\centering
	\subfloat[Biological Process]
	{\includegraphics[width=3.5in]{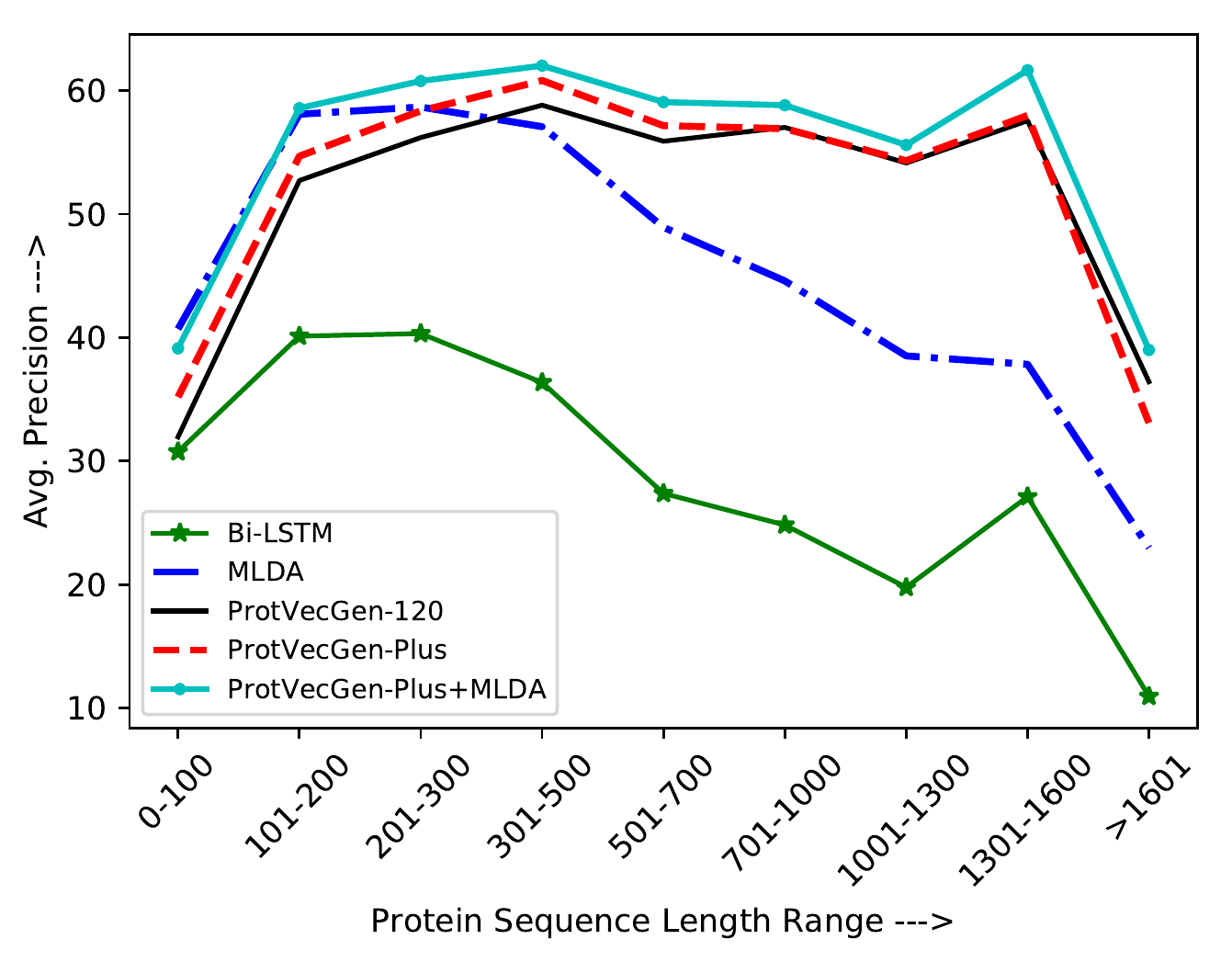}%
		\label{precision_BP}}
	\hfil
	\subfloat[Molecular Function]
	{\includegraphics[width=3.5in]{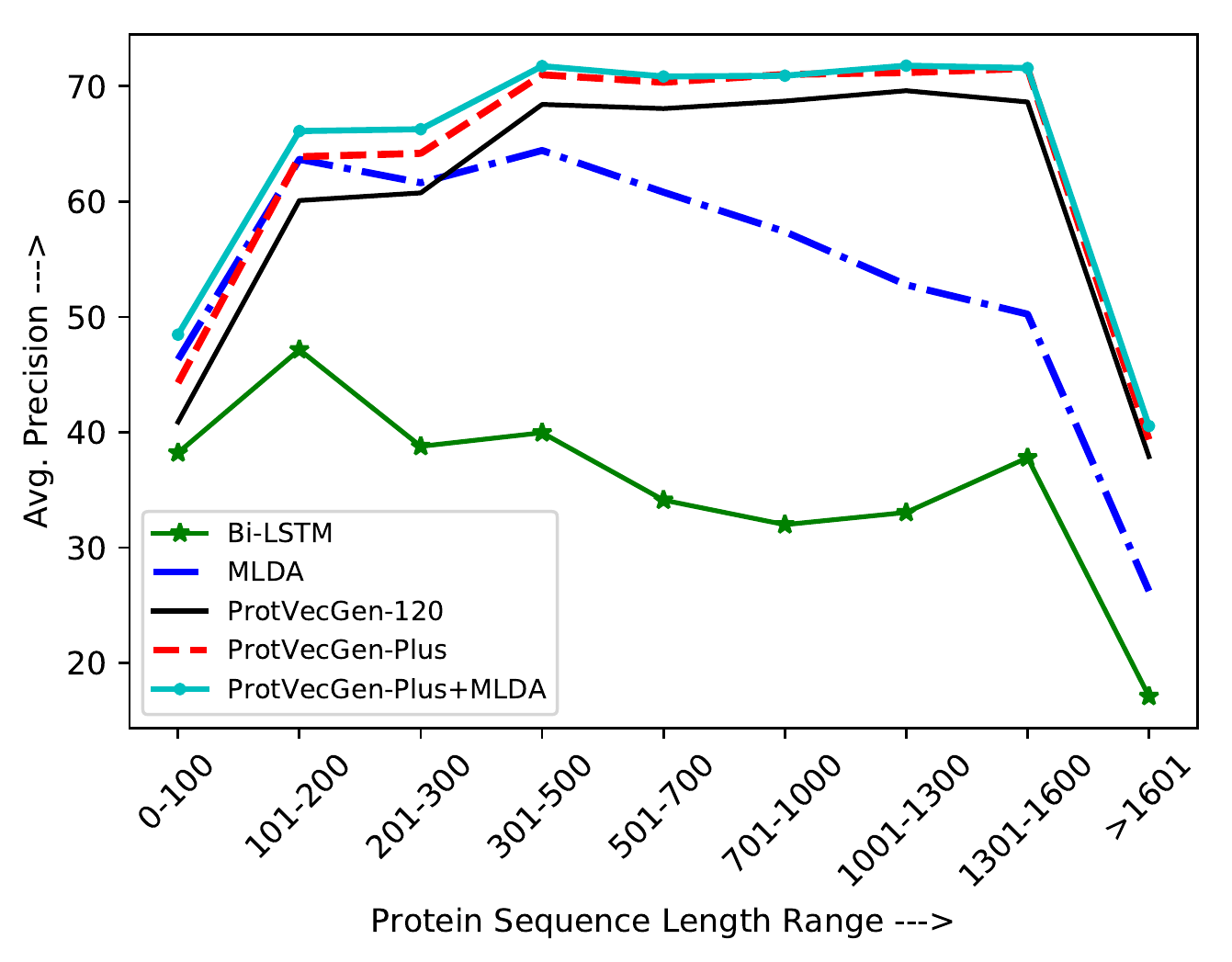}%
		\label{precision_MF}}
	\caption{Average precision for biological process and molecular function.}
	\label{Precision}
\end{figure*}

\begin{figure*}[!t]
	\centering
	\subfloat[Biological Process]
	{\includegraphics[width=3.5in]{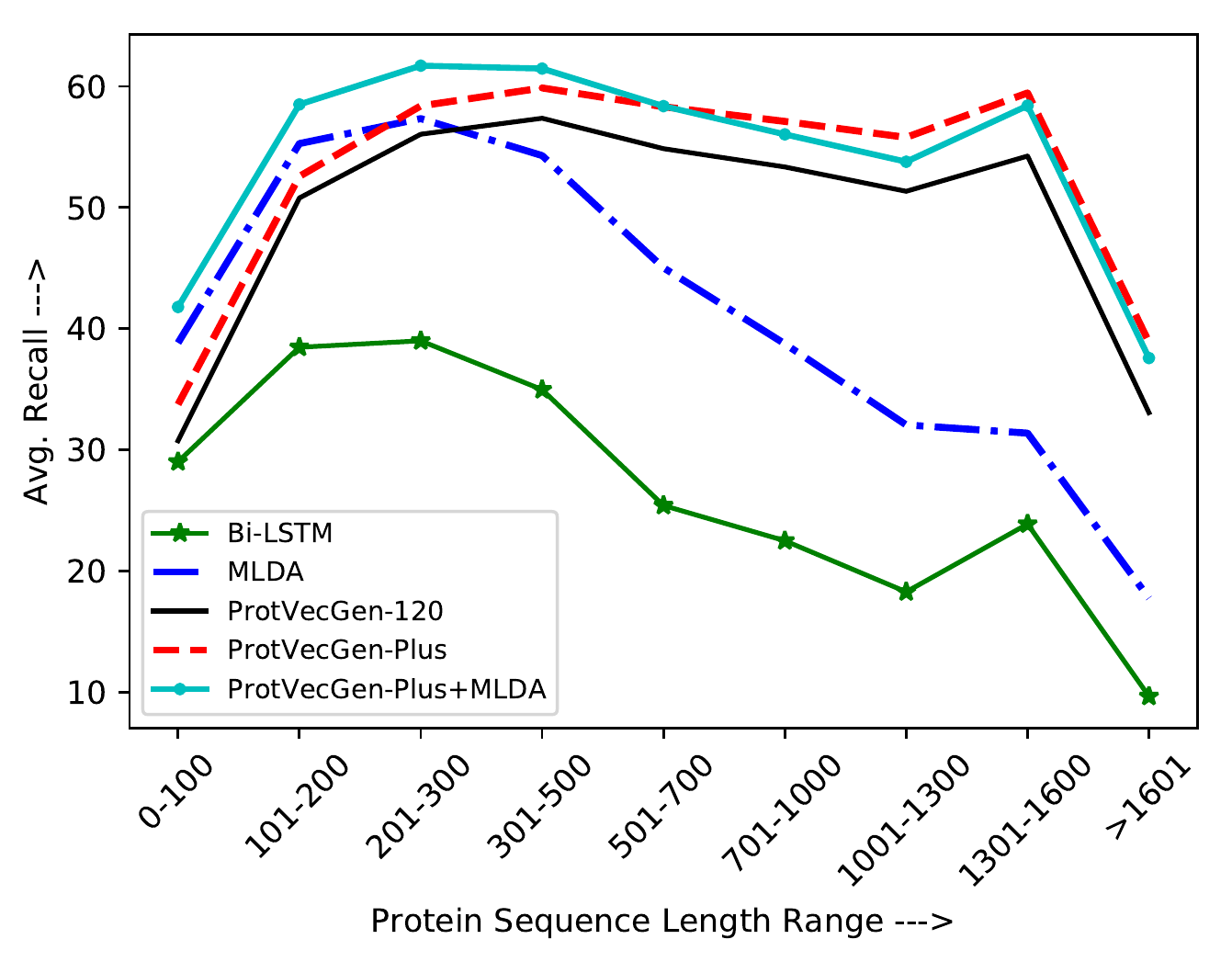}%
		\label{recall_BP}}
	\hfil
	\subfloat[Molecular Function]
	{\includegraphics[width=3.5in]{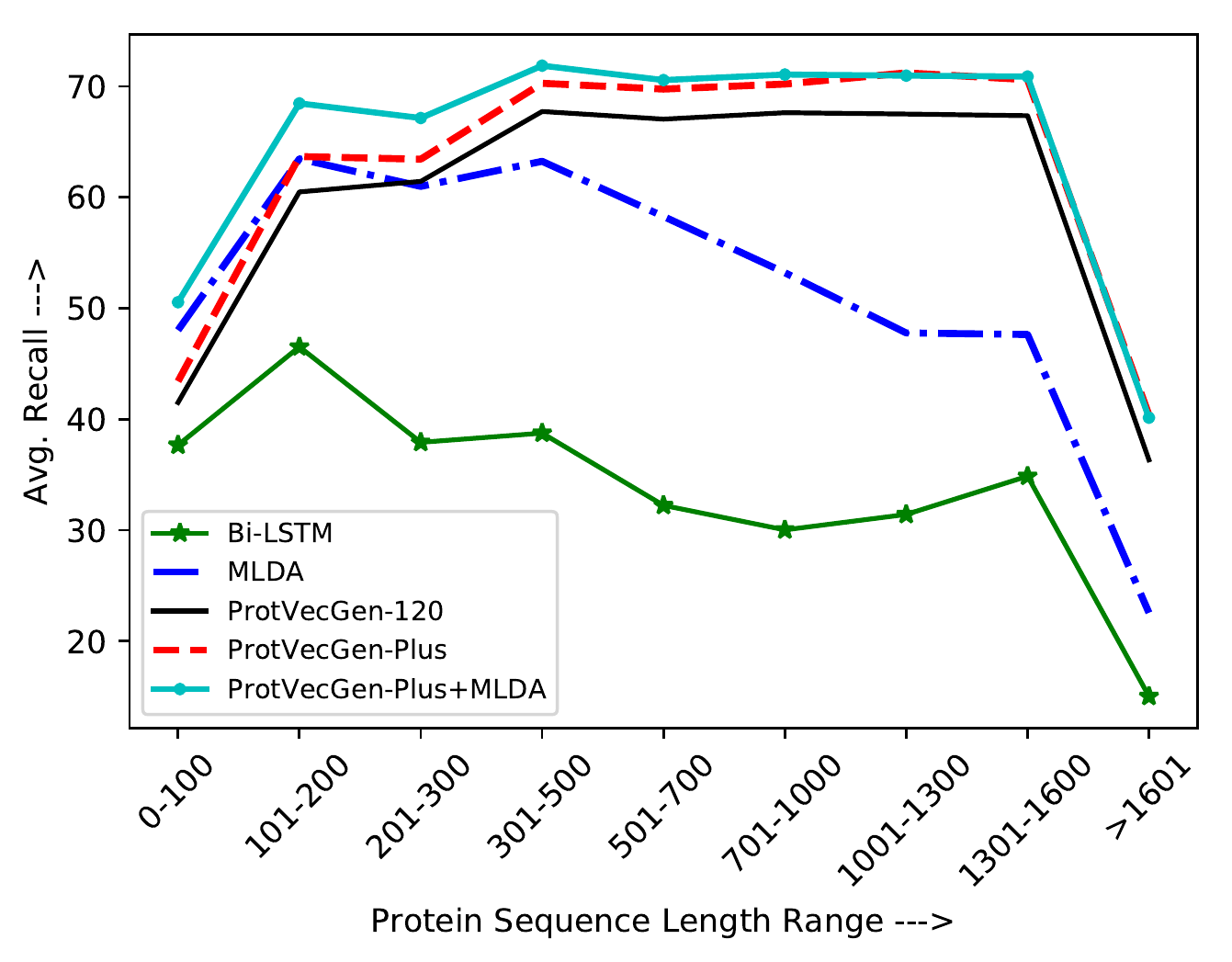}%
		\label{recall_MF}}
	\caption{Average recall for biological process and molecular function.}
	\label{Recall}
\end{figure*}

In this subsection, we investigate how the proposed feature sets perform for protein sequences of different lengths.  We choose to evaluate the performance of a feature set over nine distinct ranges of sequence lengths as shown in Fig. \ref{F1-score}-\ref{Recall}. The two proposed feature sets \textit{ProtVecGen}-120 and \textit{ProtVecGen-Plus}, along with MLDA \cite{MLDA_protFun}  and Bi-LSTM features \cite{remote_homolgy,ProtDet-CCH} (both based on analysis of complete protein sequence) from literature are used for performance evaluation. A neural network based classifier is used  as model. In addition to these, the performance of the proposed hybrid \textit{ProtVecGen-Plus}+MLDA approach is also compared with others.

Figure \ref{seq_dist_BP} and \ref{seq_dist_MF} show  the distribution of protein sequences based on sequence length for BP and MF respectively. While the majority of protein sequences are of short-to-medium length, very few are of long length. Performance of \textit{ProtVecGen-Plus} is better compared to \textit{ProtVecGen-}120 for both BP and MF as shown in Figures \ref{F1-score} (average \textit{F1}-score), \ref{Precision} (average precision values) and \ref{Recall} (average recall values). Here, MLDA is better than \textit{ProtVecGen-Plus} for protein sequences having sequence length less then 300 amino acid residues for BP. However, the performances of both \textit{ProtVecGen-Plus} and  \textit{ProtVecGen-120} features are better for sequences of having more than 300 residues when compared to MLDA. The same is true for the case of MF also with MLDA performing better for sequences with less than 200 amino acid residues and vice-versa.  

The performance of Bi-LSTM features based on complete protein sequence is worst (due to presence of high amount of \textit{noisy} regions), while that of \textit{ProtVecGen-Plus+MLDA} is superior compared to others. This is shown in  Figures \ref{F1-score}, \ref{Precision} and \ref{Recall}, which illustrate average \textit{F1}-score, average precision and average recall values respectively for both BP and MF datasets. Thus, all the three proposed approaches \textit{ProtVecGen-120}, \textit{ProtVecGen-Plus}  and \textit{ProtVecGen-Plus+MLDA} are able to mitigate the ill-effect of the uneven distribution of sequence lengths. Specifically, our proposed features showed much superior results for longer protein sequences when compared to other works in existing literature.

However, for very short ($<$ 100 amino acids) and very long ($>$ 1600 amino acids) sequences, none of the methods including proposed approach perform well as can be seen from Figures \ref{heat_map}, \ref{F1-score}, \ref{Precision} and \ref{Recall}. This is because very short protein sequences are often smaller in length than the ideal segment size (which is 100 for molecular function and 120 for biological process). In such cases, the sequences are padded up to the ideal segment size to produce the only segment. This padding in the only segment acts as noise during training and testing, adversely affecting the accuracy. Even if the sequences are a little longer than the ideal segment size, they produce very few segments – making it harder for the LSTM network to learn these sequences due to a lack of sufficient number of samples.  

Regarding the longer protein sequences, the high-proportion of the non-conserved part is responsible for producing the majority of segments. When averaging the segment vectors during creation of the global protein vectors, this majority of segments produced from non-conserved regions marginalize the useful segments produced from conserved regions. This adversely affects the learning process of the LSTM network. Removal of few irrelevant segments before creating global protein vector may help to produce much efficient protein vector, which is a major future challenge.

\subsection{Overall Comparision with State-of-The-Art and Other Approaches}
In the previous section we discussed the performance of the proposed feature sets with respect to sequence lengths. In this section, the overall performance of the proposed features is compared with features from existing literature. For evaluation, all the features are trained on the same neural network structure followed by SVM model as described in Sec. \ref{Query_model}. A neural network is composed of a single hidden layer (with activation function \textit{relu}) which is fully connected with an output layer is used as a training model. The output layer has a \textit{sigmoid} activation function as we are dealing with multi-label classification problem. Binary cross-entropy is used as a loss function.  The comparison is done for both biological process (BP) and molecular function (MF) predictions. Apart from state-of-the-art Multi-label LDA (MLDA) \cite{MLDA_protFun} features (using \textit{n}-mer with \textit{n} = 3), the features from existing literature, such as Bi-LSTM feature from Bi-directional LSTM network as in \cite{remote_homolgy,ProtDet-CCH}  (based on the analysis of complete protein sequence) and the positional statistics based Sequence Graph Transform (SGT) \cite{SGT} features (with hyper-parameters kappa($\kappa$) = 5) are also used for comparison.

Clearly, for BP, ProtVecGen-Plus outperforms MLDA and ProtVecGen-120 with an improvement of +5.38\% and +2.01\% respectively as given in Table \ref{acc_bio}. Similar results were obtained for MF, with ProtVecGen-Plus showing an improvement of +8.00\% and +2.52\% over MLDA and ProtVecGen-100 respectively. The performances of Bi-LSTM features, based on the analysis of entire sequences and SGT features, are very low compared to state-of-the-art and other proposed features. The results clearly demonstrate that the predictive performances of all the segment-based proposed approaches are significantly better than the rest. This highlights the superiority of the sub-sequence analysis approach as compared to entire sequence analysis.

The hybrid \textit{ProtVecGen-Plus+MLDA} approach does well for both biological process and molecular function. For the case of BP, average \textit{F1}-score of the hybrid approach is 56.68 $\pm$ 0.13, which is an improvement of +7.41\% and +2.03\% over the MLDA and \textit{ProtVecGen-Plus} approach respectively. Recall and precision values also exhibit similar trend as shown in Table \ref{acc_bio}. The corresponding numbers for MF prediction are: \textit{ProtVecGen-Plus+MLDA} scores 67.12 $\pm$ 0.10 on average \textit{F1} metric. This produces an improvement of +9.21\% and +1.21\% over the MLDA and \textit{ProtVecGen-Plus} approach respectively. Recall and precision values also exhibit similar trend. Overall, \textit{ProtVecGen-Plus+MLDA} feature outperforms all other feature sets and does well while annotating new proteins.

\section{Conclusion}\label{conclusion}
The proposed framework generates a highly discriminative feature vector corresponding to a protein sequence by splitting the sequence into smaller segments, which are eventually used for constructing feature vectors.  The approach overcomes the side-effects associated with using a protein sequence in entirety for constructing feature vectors by reducing the effect of \textit{noisy} or \textit{non-conserved} regions over the \textit{conserved} pattern.

The proposed approach also takes care of the variable size of conserved patterns by converting a protein sequence into a vector based on segments of multiple sizes - 100, 120, and 140. The new framework greatly improves the function prediction performance over the existing techniques and also shows consistent performance for protein sequences of different lengths.

The proposed framework is also resistant to the uneven distribution of protein sequences based on sequence length. Further, the proposed approach is combined with MLDA feature based model to improve the overall performance using the new weighted scheme based on individual average \textit{F1}-score of each model. Future work will be to incorporate interaction data to improve upon the performance for the biological processes and handle a large number of $GO$-terms.

\section*{Acknowledgments}
Akshay Deepak has been awarded Young Faculty Research Fellowship (YFRF) of Visvesvaraya PhD Programme of Ministry of Electronics \& Information Technology, MeitY, Government of India. In this regard, he would like to acknowledge that this publication is an outcome of the R\&D work undertaken in the project under the Visvesvaraya PhD Scheme of Ministry of Electronics \& Information Technology, Government of India, being implemented by Digital India Corporation (formerly Media Lab Asia).

\ifCLASSOPTIONcaptionsoff
  \newpage
\fi



%

\begin{IEEEbiography}[{\includegraphics[width=0.9in,height=1.3in,clip,keepaspectratio]{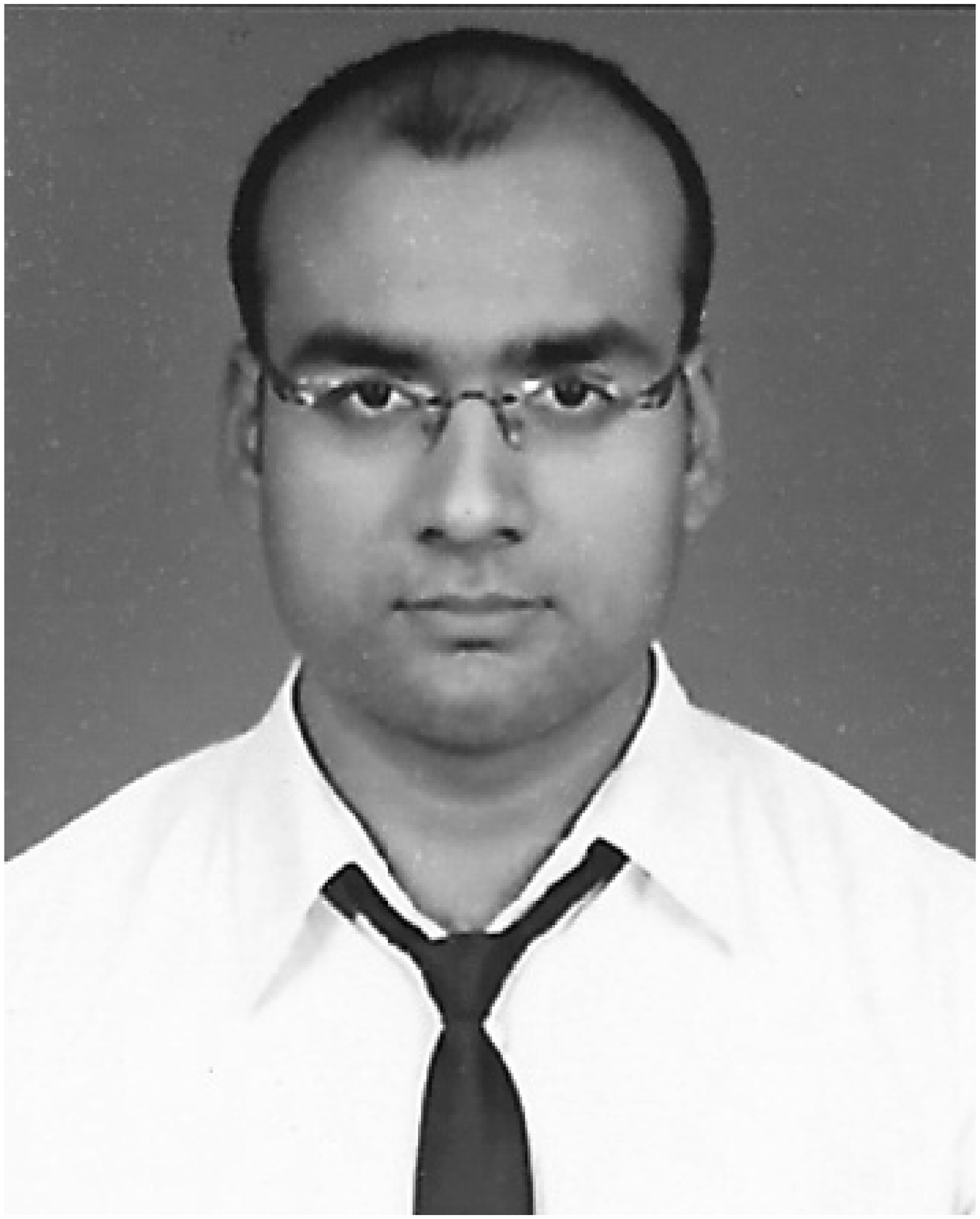}}]{Ashish Ranjan} is currently pursuing Ph.D. in Computer Science and Engineering from National Institute of Technology Patna, India. His research interest include machine learning, artificial neural network, and computational biology.
\end{IEEEbiography}
\begin{IEEEbiography}[{\includegraphics[width=0.9in,height=1.3in,clip,keepaspectratio]{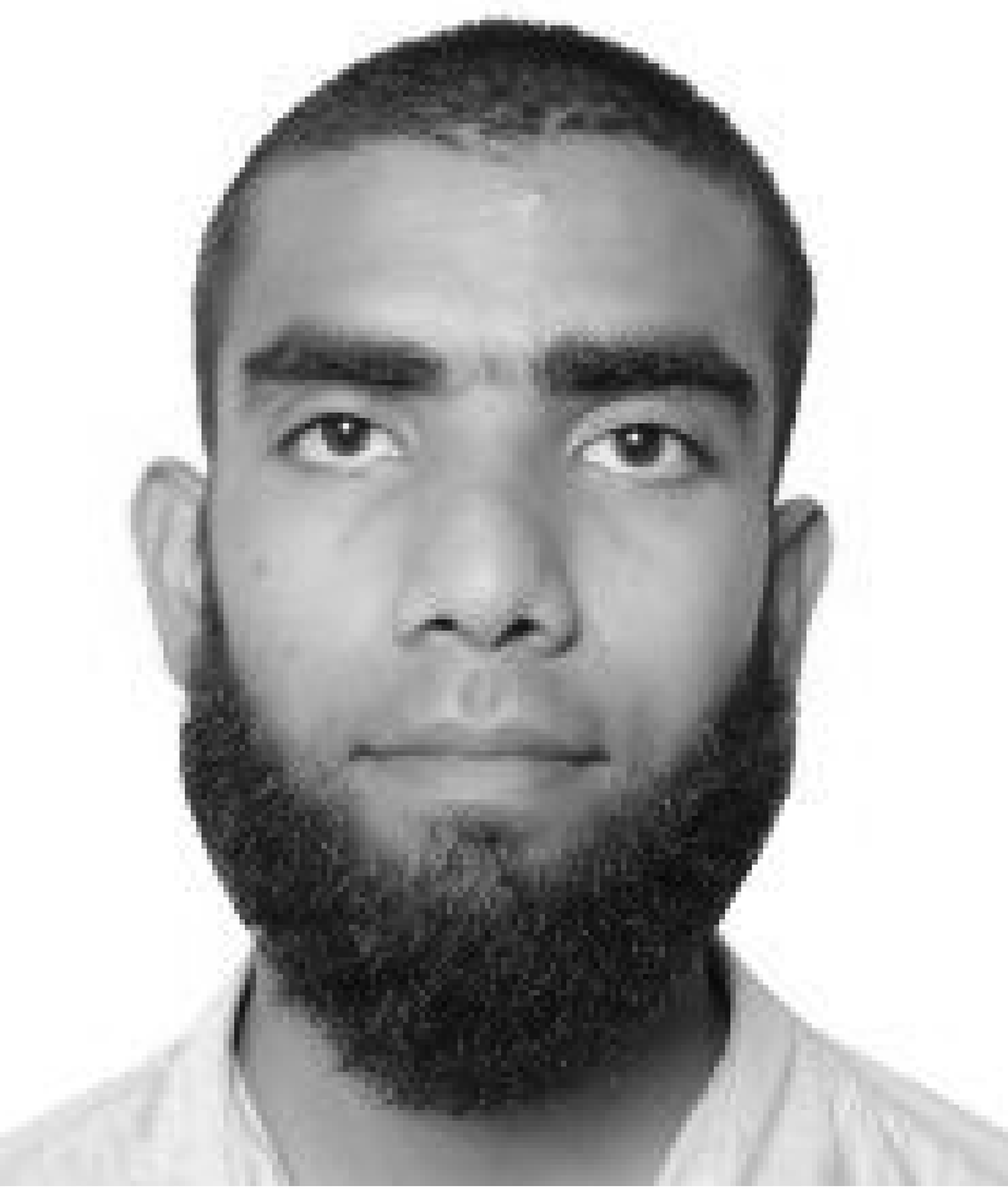}}]{Md Shah Fahad} is currently pursuing Ph.D. in Computer Science and Engineering from National Institute of Technology Patna, India. His research interest include machine learning, speech processing and pattern recognition.
\end{IEEEbiography}

\begin{IEEEbiography}[{\includegraphics[width=0.9in,height=1.3in,clip,keepaspectratio]{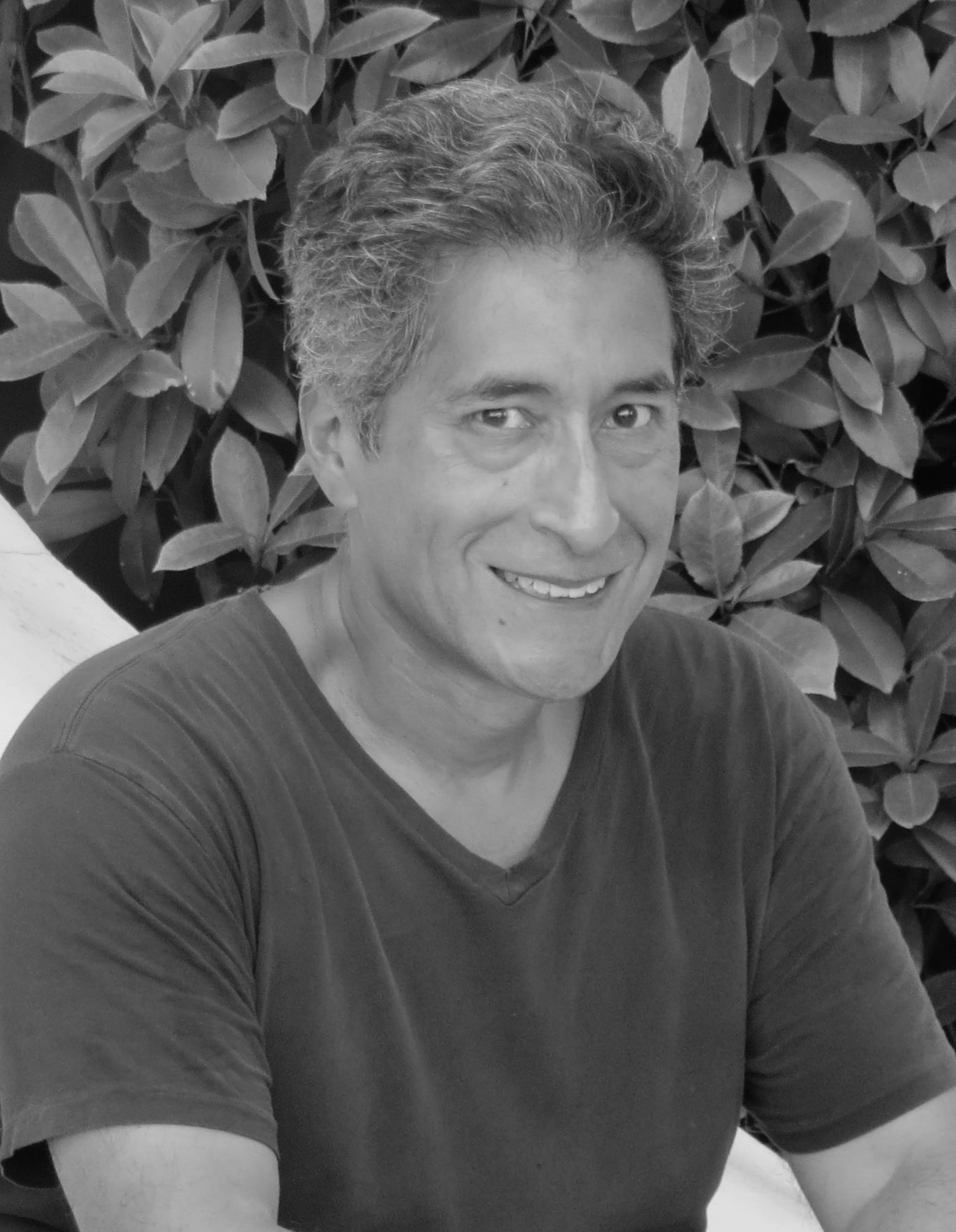}}]{David Fern$\acute{\textbf{a}}$ndez-Baca} is a professor in the Department of Computer Science, Iowa State University, USA. His research interest includes combinatorial algorithms, phylogenetic tree construction, computational biology, and sensitivity analysis of optimization problems. \url{http://web.cs.iastate.edu/~fernande/}
\end{IEEEbiography}

\begin{IEEEbiography}[{\includegraphics[width=0.9in,height=1.3in,clip,keepaspectratio]{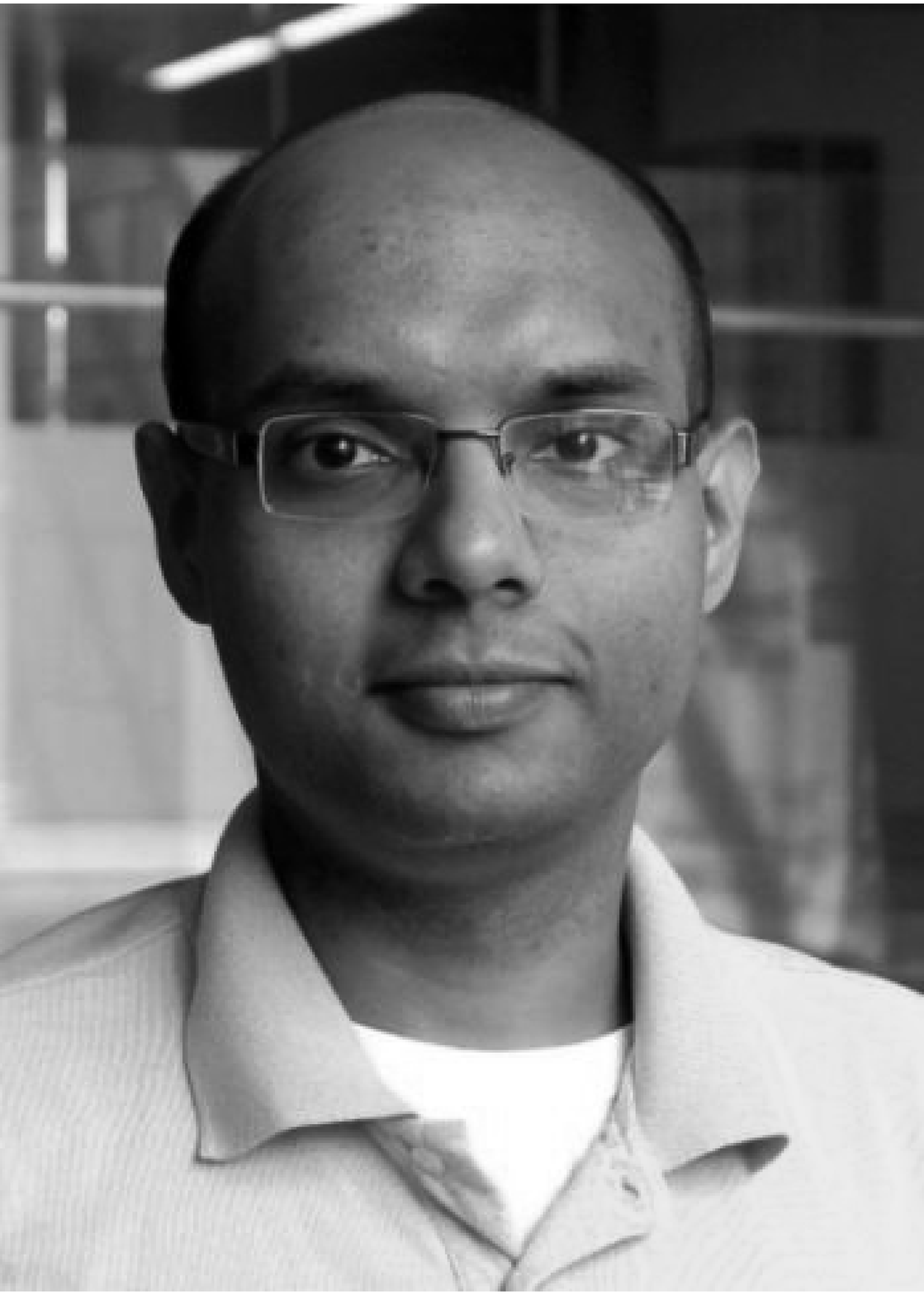}}]{Akshay Deepak} is an assistant professor in the Department of Computer Science and Engineering, National Institute of Technology Patna, India. He received his MS and PhD degrees in Computer Science from Iowa State University, USA. His research interest include data mining, machine learning and computational biology. \url{http://www.nitp.ac.in/php/faculty_profile.php?id=akshayd@nitp.ac.in} 
\end{IEEEbiography}

\begin{IEEEbiography}[{\includegraphics[width=1in,height=1.1in,clip]{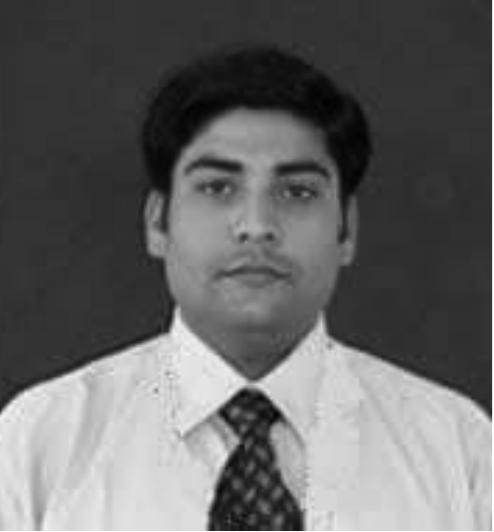}}]{Sudhakar Tripathi} is an associate professor in the Department of Information Technology, REC Ambedkar Nagar, UP, India. He received his PhD degree in Computer Science from Indian Institute of Technology (BHU) Varanasi, India. His research interest include data mining, artificial neural network  and computational biology.
\end{IEEEbiography}


\vfill

\end{document}